\documentclass[acmsmall,screen,final]{acmart}

\AtBeginDocument{%
  }

\setcopyright{none}

\acmJournal{TELO}
\acmVolume{0}
\acmNumber{0}
\acmArticle{0}

\usepackage{latexsym}
\usepackage{amsmath} 
\usepackage{mathtools}       
\usepackage{stmaryrd}        
\usepackage{bm}              
\DeclareMathOperator{\Tr}{Tr}     
\DeclareMathOperator{\Cov}{Cov}   
\DeclareMathOperator{\diag}{diag} 
\DeclareMathOperator{\vect}{vec}  

\providecommand{\E}{\mathbb{E}} 
\providecommand{\T}{\mathrm{T}} 
\renewcommand{\leq}{\leqslant} 
\DeclarePairedDelimiterX{\inner}[2]{\langle}{\rangle}{#1, #2}
\DeclarePairedDelimiter{\norm}{\lVert}{\rVert}

\usepackage{amsthm}
\usepackage{xcolor}
\usepackage{ifthen}
\usepackage{cleveref}

\theoremstyle{definition}

\usepackage{algorithm}
\usepackage{algpseudocode}

\algnewcommand\algorithmicset{\textbf{Set:}}
\algnewcommand\Set{\item[\algorithmicset]}

\providecommand{\mvae}{\mathcal{E}}
\providecommand{\mvav}{\mathcal{V}}
\providecommand{\etam}{\eta_{m}}
\providecommand{\etasig}{\eta_{\Sigma}}

\makeatletter
\let\OldStatex\Statex
\renewcommand{\Statex}[1][3]{%
  \setlength\@tempdima{\algorithmicindent}%
  \OldStatex\hskip\dimexpr#1\@tempdima\relax}
\makeatother

\definecolor{yelloworange}{RGB}{255, 174, 66}

\begin{document}

\title{CMA-ES with Learning Rate Adaptation}

\author{Masahiro Nomura}
\authornote{Corresponding author}
\email{nomura.m.ad@m.titech.ac.jp}
\orcid{0000-0002-4945-5984}
\affiliation{%
  \institution{Tokyo Institute of Technology}
  \streetaddress{4259 Nagatsutach\={o}, Midori Ward}
  \city{Yokohama}
  \state{Kanagawa}
  \country{Japan}
  \postcode{226-0026}
}
\author{Youhei Akimoto}
\email{akimoto@cs.tsukuba.ac.jp}
\orcid{0000-0003-2760-8123}
\affiliation{%
  \institution{University of Tsukuba \& RIKEN AIP}
  \streetaddress{1-1-1 Tennodai}
  \city{Tsukuba}
  \state{Ibaraki}
  \country{Japan}
  \postcode{305-8573}
}

\author{Isao Ono}
\email{isao@c.titech.ac.jp}
\orcid{0009-0008-2110-9853}
\affiliation{%
  \institution{Tokyo Institute of Technology}
  \streetaddress{4259 Nagatsutach\={o}, Midori Ward}
  \city{Yokohama}
  \state{Kanagawa}
  \country{Japan}
  \postcode{226-0026}
}

\renewcommand{\shortauthors}{Nomura et al.}

\begin{abstract}
The covariance matrix adaptation evolution strategy (CMA-ES) is one of the most successful methods for solving continuous black-box optimization problems.
A practically useful aspect of CMA-ES is that it can be used without hyperparameter tuning.
However, the hyperparameter settings still have a considerable impact on performance, especially for \emph{difficult} tasks, such as solving multimodal or noisy problems.
This study comprehensively explores the impact of learning rate on CMA-ES performance and demonstrates the necessity of a small learning rate by considering ordinary differential equations.
Thereafter, it discusses the setting of an ideal learning rate.
Based on these discussions, we develop a novel learning rate adaptation mechanism for CMA-ES that maintains a constant signal-to-noise ratio.
Additionally, we investigate the behavior of CMA-ES with the proposed learning rate adaptation mechanism through numerical experiments and compare the results with those obtained for CMA-ES with a fixed learning rate and with population size adaptation.
The results show that CMA-ES with the proposed learning rate adaptation works well for multimodal and/or noisy problems \emph{without} extremely expensive learning rate tuning.
\end{abstract}




\setcopyright{acmlicensed}
\acmJournal{TELO}
\acmYear{2024} \acmVolume{1} \acmNumber{1} \acmArticle{1} \acmMonth{1}\acmDOI{10.1145/3698203}

\maketitle

\section{Introduction}
\label{sec:intro}
The covariance matrix adaptation evolution strategy (CMA-ES)~\cite{hansen2001completely,hansen2016cma} is among the most successful methods available for solving continuous black-box optimization problems;
its effectiveness has been confirmed through various real-world applications~\cite{nomura2021warm,kikuchi2021constrained,kikuchi2021modeling,maki2020application,fujii2018exploring,ha2018world,volz2018evolving,tanabe2021level,huang2022action,piergiovanni2020evolving,tian2023multi,purucker2023cma}.
CMA-ES performs optimization by updating the multivariate Gaussian distribution; that is, it first samples candidate solutions from the distribution and then updates the distribution parameters (i.e., the mean vector $m$ and covariance matrix $\Sigma = \sigma^2 C$) based on the objective function $f$.
This update is partly based on the natural gradient descent~\cite{akimoto2010bidirectional,ollivier2017information} of the expected $f$, and
$m$ and $C$ in CMA-ES are updated to reduce the expected evaluation value.
CMA-ES is practically useful as it is a quasi-hyperparameter-free algorithm;
practitioners can use it without hyperparameter tuning because default values are provided for all hyperparameters through theoretical analysis and extensive empirical evaluations.
Specifically, the hyperparameter values are automatically computed using dimension $d$ and population size $\lambda$, where $\lambda = 4 + \lfloor 3 \ln(d) \rfloor$ by default.

Although the default $\lambda$ value works well for various unimodal problems, increasing it can help solve \emph{difficult} tasks, such as solving multimodal and additive noise problems~\cite{hansen2004evaluating,psaigo,psacmaes}.
However, in a black-box scenario, determining the problem structure of $f$ is challenging.
Thus, determining the appropriate $\lambda$ value in advance is also challenging, and online adaptation of $\lambda$ has been proposed to address the issue~\cite{psaigo,psacmaes,pccmsaes,cmaesapop}.
Population size adaptation (PSA)-CMA-ES~\cite{psacmaes} is a representative $\lambda$ adaptation mechanism that has exhibited promising performance for difficult tasks, including multimodal and additive noise problems.

It has been observed that, in CMA-ES, increasing $\lambda$ has an effect similar to decreasing the $m$ learning rate, that is, $\eta_m$~\cite{cmasmallcm}\footnote{Note that, in Ref.~\cite{cmasmallcm}, the rank-one update was excluded from CMA-ES. In this study, however, we consider CMA-ES including the rank-one update.}.
Indeed, the $m$ and $\Sigma$ learning rates, that is, $\eta$, is another hyperparameter that critically affects performance.
An excessively large $\eta$ value results in unstable parameter updates, whereas an excessively small value degrades search efficiency.
Miyazawa and Akimoto~\cite{cmasmallcm} reported that CMA-ES with even a relatively small $\lambda$ (e.g., $\lambda = \sqrt{d}$) solves multimodal problems through an appropriate setting of $\eta$.
However, determining the appropriate $\eta$ value is difficult in practice because prior knowledge is often limited and hyperparameter tuning entails expensive numerical investigations.

Therefore, online adaptation of $\eta$ based on the problem difficulty constitutes an important advancement as it will allow practitioners to \emph{safely} use CMA-ES without requiring prior knowledge or expensive trial-and-error calculations.
In particular, we believe that $\eta$ adaptation is more advantageous than $\lambda$ adaptation from a practical perspective because the former is more suitable for parallel implementations.
For example, practitioners often wish to specify a certain number of workers as the value of $\lambda$ value to avoid wasting computational resources.
However, $\lambda$ adaptation may not always effectively utilize the available resources, as the values vary during the optimization process.
By contrast, $\eta$ adaptation allows complete exploitation of the available resources because the value of $\lambda$ is fixed as the maximum number of workers.
Moreover, in $\eta$ adaptation, the parameters are regularly updated, whereas CMA-ES with $\lambda$ adaptation does not progress until all $\lambda$ solutions are evaluated, making it difficult to determine the search termination point.

Although online $\eta$ adaptation itself is not new and several studies have attempted to adapt $\eta$ values in CMA-ES variants, these adaptations targeted \emph{speed-up}~\cite{lradaptnes,gissler2022learning,loshchilov2014maximum}.
One notable exception is the $\eta$ adaptation proposed by Krause~\cite{noiseresilientes} that aims to solve additive noise problems through new evolution strategies.
However, it estimates the problem difficulty through resampling, that is, by repeatedly evaluating the \emph{same} solution; thus, it is not suitable for solving (noiseless) multimodal problems.
Furthermore, as it involves significant modifications of the internal parameters of the evolution strategies, applying it directly to CMA-ES is challenging.

This study aimed to develop CMA-ES to solve multimodal and additive noise problems without extremely expensive $\eta$ tuning or adjusting any other CMA-ES parameters except $\eta$.
To achieve this, we first examined the impact of learning rate.
Our results suggest that (i) difficult problems can be relatively easily solved by decreasing the learning rate and aligning the parameter behavior with the trajectory of an ordinary differential equation (ODE), and (ii) the optimal learning rate is approximately proportional to the signal-to-noise ratio (SNR).
Based on these observations, we propose an $\eta$ adaptation mechanism for CMA-ES---called the learning rate adaptation (LRA)---that adapts $\eta$ to maintain a constant SNR.
The key feature of the proposed method is that it does not require specific knowledge of the internal mechanism of the distribution-parameter update to estimate the SNR.
Consequently, the proposed method is widely applicable to various CMA-ES variants, such as diagonal decoding (dd)-CMA~\cite{ddcma}, even though this study considers the most commonly used CMA-ES, which combines weighted recombination, step-size $\sigma$ adaptation, rank-one update, and rank-$\mu$ update.

It should be noted that our work focuses on well-structured multimodal problems rather than weakly structured ones, as in the previous studies on $\lambda$ adaptation in CMA-ES~\cite{psacmaes}.
Using our method alone cannot solve the weakly structured multimodal problems and may even be detrimental to these problems.
To address such problems, we believe the integration of restart strategies (e.g., BIPOP-CMA-ES~\cite{hansen2009benchmarking}) is necessary, which is beyond this study; thus, we have left it for future work.

This study extends a previous study~\cite{nomura2023cma} as follows:
In Section~\ref{sec:decreasing_lr}, we illustrate from an ODE perspective the reason why a small learning rate is essential for solving difficult problems.
Thereafter, we discuss the optimal learning rate in Section~\ref{sec:optimal_lr}, which indicates that the proposed method adapts the learning rate to a nearly optimal value.
It should be noted that Section~\ref{sec:closer_look} presents entirely new information that was not included in the previous study~\cite{nomura2023cma}.
Thereafter, the performance differences for various $\lambda$ values are discussed in Section~\ref{sec:exp_popsize}.
Finally, Section ~\ref{sec:exp_vs_psa} presents a comprehensive comparison of LRA-CMA-ES and PSA-CMA-ES~\cite{psacmaes}, a state-of-the-art $\lambda$ adaptation method.

The remainder of this paper is organized as follows:
Section~\ref{sec:background} explains the CMA-ES algorithm and the information-geometric optimization (IGO) framework.
Section~\ref{sec:closer_look} closely examines and explains the impact of the learning rate and presents the discussion for determining the ideal learning rate.
Section~\ref{sec:lra} presents the proposed $\eta$ adaptation mechanism based on SNR estimation.
Section~\ref{sec:exp} evaluates the performance of the proposed $\eta$ adaptation for noiseless and noisy problems.
Finally, Section~\ref{sec:conclusion} concludes the paper and suggests future research directions.

\section{Background}
\label{sec:background}

\subsection{CMA-ES}
\label{sec:cma}

We consider minimizing the objective function $f: \mathbb{R}^d \to \mathbb{R}$.
CMA-ES employs a multivariate Gaussian distribution to generate candidate solutions, where the distribution $\mathcal{N}(m, \sigma^2 C)$ is parameterized through three elements: mean vector $m \in \mathbb{R}^d$, step-size $\sigma \in \mathbb{R}_{>0}$, and covariance matrix $C \in \mathbb{R}^{d\times d}$.

CMA-ES first initializes the $m^{(0)}, \sigma^{(0)}$, and $C^{(0)}$ parameters.
Thereafter, the following steps are repeated until a predefined stopping criterion is met.

\noindent
{\bf Step 1. Sampling and Evaluation}\\
At iteration $t + 1$ (where $t$ begins at $0$), $\lambda$ candidate solutions $x_i\ (i=1, 2, \cdots, \lambda)$ are sampled independently from $\mathcal{N}(m^{(t)}, ( \sigma^{(t)} )^2 C^{(t)})$, as follows:
\begin{align}
    y_i &= \sqrt{ C^{(t)} } z_i, \label{eq:y}\\
    x_i &= m^{(t)} + \sigma^{(t)} y_i, \label{eq:x}
\end{align}
where $z_i \sim \mathcal{N}(0, I)$ and $I$ is the identity matrix.
The solutions are evaluated on $f$ and sorted in ascending order.
Let $x_{i:\lambda}$ be the $i$-th best candidate solution, that is, $f(x_{1:\lambda}) \leq f(x_{2:\lambda}) \leq \cdots \leq f(x_{\lambda:\lambda})$ for minimization.
In addition, we let $y_{i:\lambda}$ and $z_{i:\lambda}$ be the intermediate vectors in \Cref{eq:y,eq:x} corresponding to $x_{i:\lambda}$.

\noindent
{\bf Step 2. Compute Evolution Paths}\\
The weighted averages $dy = \sum_{i=1}^{\mu} w_i y_{i:\lambda}$ and $dz = \sum_{i=1}^{\mu} w_i z_{i:\lambda}$ of the intermediate vectors are calculated using the parent number $\mu \leq \lambda$ and weight function $w_i$, where $\sum_{i=1}^{\mu} w_i=1$.
The evolution paths are updated as follows:
\begin{align}
    p_{\sigma}^{(t+1)} &= (1-c_{\sigma}) p_{\sigma}^{(t)} + \sqrt{c_{\sigma} (2-c_{\sigma}) \mu_w} dz, \\
    p_{c}^{(t+1)} &= (1-c_{c}) p_{c}^{(t)} + h_{\sigma}^{(t+1)} \sqrt{c_{c} (2-c_{c}) \mu_w} dy,
\end{align}
where $\mu_w = 1 / \sum_{i=1}^{\mu} w_i^2$, $c_{\sigma}$, and $c_c$ are the cumulation factors, and $h_{\sigma}^{(t+1)}$ is the Heaviside function, which is defined as follows~\cite{hansen2014principled}:
\begin{align}
    h_{\sigma}^{(t+1)} = \begin{cases}
1 & {\rm if}\ \frac{\| p_{\sigma}^{(t+1)} \|^2}{1 - (1-c_{\sigma})^{2(t+1)}} < \left( 2 + \frac{4}{d+1} \right) d, \\
0 & {\rm otherwise}.
\end{cases}
\end{align}

\noindent
{\bf Step 3. Update Distribution Parameters}\\
The distribution parameters are updated as follows~\cite{hansen2014principled}:
\begin{align}
    \label{eq:m}
    &m^{(t+1)} = m^{(t)} + c_m \sigma^{(t)} dy, \\
    \label{eq:csa}
    &\sigma^{(t+1)} = \sigma^{(t)} \exp \left( \min \left(1, \frac{c_{\sigma}}{d_{\sigma}} \left( \frac{\| p_{\sigma}^{(t+1)} \|}{\mathbb{E}[\| \mathcal{N}(0, I) \|]} - 1 \right) \right) \right), \\
    \label{eq:C}
    &\begin{multlined}[t]C^{(t+1)} = \left( 1 + (1-h_{\sigma}^{(t+1)}) c_1 c_c (2-c_c) \right) C^{(t)} \\
    +\underbrace{c_1 \left[ p_{c}^{(t+1)} \left( p_{c}^{(t+1)} \right)^{\top} - C^{(t)} \right]}_{\text{rank-one update}} + \underbrace{c_{\mu} \sum_{i=1}^{\mu} w_i \left[ y_{i:\lambda} y_{i:\lambda}^{\top} - C^{(t)} \right]}_{\text{rank-$\mu$ update}},\end{multlined}
\end{align}
where $\mathbb{E}[\| \mathcal{N}(0, I) \|] \approx \sqrt{d} \left( 1 - \frac{1}{4d} + \frac{1}{21 d^2} \right)$ denotes the expected Euclidean norm of the sample of a standard normal distribution and $c_m$ is the learning rate for $m$, which is typically set to $1$.
$c_1$ and $c_{\mu}$ are the learning rates for the rank-one and -$\mu$ updates of $C$, respectively, and
$d_{\sigma}$ is the damping factor for the $\sigma$ adaptation.

\subsection{IGO}
\label{sec:igo}
The IGO~\cite{ollivier2017information} framework is a unified framework for stochastic search methods.
Given a family of probability distributions parameterized by $\theta \in \Theta$, the original objective function $f$ is transformed into a new objective function $J_{\theta}$ that is defined in the distribution-parameter space $\Theta$.

For the family of Gaussian distributions, the IGO algorithms recover the pure rank-$\mu$-update CMA-ES, eliminating the $\sigma$ adaptation and rank-one update from the procedures in \Cref{sec:cma}. To investigate the effects of learning rates on CMA-ES, we focus on their properties within the context of the IGO framework with a family of Gaussian distributions in \Cref{sec:closer_look}.
This section presents the background of the IGO framework.

Instead of minimizing the original objective $f$ over the input domain $\mathbb{R}^d$, IGO maximizes a new objective $J_{\theta}$ over the distribution-parameter domain $\Theta$.
Let $u: [0, 1] \to \mathbb{R}$ be a bounded, non-increasing function, and $P_{\theta}$ be the Lebesgue measure on $\mathbb{R}^d$ corresponding to the probability density $p(x; \theta)$.
We define the utility function $W_{\theta}^{f}$ as
\begin{align}
    W_{\theta}^{f}(x) = u( q_{\theta}(x) ) ,
\end{align}
where $q_{\theta}(x)$ is the quantile function that is defined as $q_{\theta}(x) := P_{\theta}[y: f(y) \leq f(x)]$ for minimization.
The objective updating $\theta$, given the \emph{current} distribution parameters $\theta^{(t)}$, is defined as the expectation of the weighted quantile function $W_{\theta^{(t)}}^{f}(x)$ over $p(x;\theta)$:
\begin{align}
    J_{\theta^{(t)}}(\theta) = \mathbb{E}_{x \sim p(x;\theta)} [W_{\theta^{(t)}}^{f}(x)] .
\end{align}

The objective $J_{\theta^{(t)}}(\theta)$ is maximized based on the \emph{natural} gradient~\cite{amari1998why,amari2000methods}.
By using the ‘‘log-likelihood trick’’ under some mild conditions, the \emph{vanilla} gradient can be calculated as
\begin{align}
    \nabla_{\theta} J_{\theta^{(t)}}(\theta) = \mathbb{E}_{x \sim p(x;\theta)} [W_{\theta^{(t)}}^{f}(x) \nabla_{\theta} \ln p(x; \theta)] .
\end{align}
The \emph{natural} gradient is obtained through the product of the inverse of the Fisher information matrix $F$ at $\theta^{(t)}$ and the vanilla gradient as follows:
\begin{align}
    \tilde{\nabla}_{\theta} J_{\theta^{(t)}}(\theta) = \mathbb{E}_{x \sim p(x;\theta)} [W_{\theta^{(t)}}^{f}(x) \tilde{\nabla}_{\theta} \ln p(x; \theta)] ,
\end{align}
where $\tilde{\nabla}_{\theta} \ln p(x; \theta) = F^{-1} \nabla_{\theta} \ln p(x; \theta)$.

In practice, the integral cannot be calculated in a closed form and is therefore estimated using the Monte Carlo method as follows:
\begin{align}
    \tilde{\nabla}_{\theta} J_{\theta^{(t)}}(\theta) \approx \frac{1}{\lambda} \sum_{i=1}^{\lambda} W_{\theta^{(t)}}^{f}(x_i) \tilde{\nabla}_{\theta} \ln p(x_i; \theta),
\end{align}
where $\{ x_i \}_{i=1}^{\lambda}$ are $\lambda$ i.i.d. samples obtained from probability distribution $p(x_i; \theta)$.
The IGO algorithms implement the IGO framework using the estimated natural gradient, whose updated equation is as follows:
\begin{align}
    \label{eq:igo_update}
    \theta^{(t+1)} = \theta^{(t)} + \eta \sum_{i=1}^{\lambda} \frac{W_{\theta^{(t)}}^{f}(x_i)}{\lambda} \tilde{\nabla}_{\theta} \ln p(x_i; \theta^{(t)}),
\end{align}
where $\eta$ denotes the learning rate.
In practice, $W_{\theta^{(t)}}^{f}(x_i)$ is also estimated based on the \emph{ranking} of $\{ x_i \}_{i=1}^{\lambda}$.

As elucidated herein, the IGO framework, with a family of Gaussian distributions, recovers the rank-$\mu$-update CMA-ES~\cite{akimoto2010bidirectional,ollivier2017information}.
If the distribution parameter $\theta = (m^{\top}, \vect(C)^{\top})^{\top}$, then ~\cite{akimoto2010bidirectional}:
\begin{align}
    \tilde{\nabla}_{\theta} \ln p(x; \theta) = 
\begin{pmatrix}
    x - m \\
    \vect( (x-m)(x-m)^{\top} - C )
\end{pmatrix}.
\end{align}
Thus, Eq.~(\ref{eq:igo_update}) can be rewritten as 
\begin{align}
    \label{eq:igo_m}
    m^{(t+1)} &= m^{(t)} + \eta \sum_{i=1}^{\lambda} \frac{W_{\theta^{(t)}}^{f}(x_i)}{\lambda} (x_i - m^{(t)}), \\
    \label{eq:igo_C}
    C^{(t+1)} &= C^{(t)} + \eta \sum_{i=1}^{\lambda} \frac{W_{\theta^{(t)}}^{f}(x_i)}{\lambda} \left( (x_i-m^{(t)})(x_i-m^{(t)})^{\top} - C^{(t)} \right) .
\end{align}
Consequently, by ignoring the $\sigma$ adaptation and rank-one update in CMA-ES, assuming $c_m = c_{\mu} (:= \eta)$, and considering that $w_i$ in CMA-ES is an approximation of $W_{\theta^{(t)}}^{f}(x_i) / \lambda$ in the IGO update,
the $m$ and $C$ updates through the IGO algorithm (Eqs.~(\ref{eq:igo_m}) and~(\ref{eq:igo_C}), respectively) align with those of CMA-ES (Eqs.~(\ref{eq:m}) and~(\ref{eq:C}), respectively).

\section{Learning Rate Impact}
\label{sec:closer_look}

In this section, we discuss the impact of the learning rate on CMA-ES.
First, Section~\ref{sec:relation_pop_lr} summarizes existing research on adjusting the population size, which is a common practice for difficult tasks, such as multimodal problems, and the relation between the population size and learning rate. 
In Section~\ref{sec:decreasing_lr}, we discuss the behavior from the perspective of ODEs for small learning rates.
Consequently, we demonstrate that difficult problems can be solved by reducing the learning rate (i.e., closer to the solution of the ODE). However, it should be noted that an excessively small learning rate can reduce the search efficiency.
Therefore, Section~\ref{sec:optimal_lr} discusses the determination of the optimal learning rate.

\subsection{Relation Between Population Size and Learning Rate}
\label{sec:relation_pop_lr}

Previous studies generally focused on increasing the population size $\lambda$ to solve multimodal problems.
\citet{hansen2004evaluating} reported that CMA-ES with a sufficiently large population size can often solve multimodal problems with high probability.
Based on this observation, \citet{auger2005restart} proposed IPOP-CMA-ES, which doubles the population size with each restart.
Although these studies considered CMA-ES with default learning rates, \citet{cmasmallcm} experimentally evaluated the performance of CMA-ES using small learning rates and showed that multimodal problems, such as the Rastrigin function, can be solved by setting sufficiently small learning rates \emph{without} using a large population size.
This empirical observation suggests that the effect of increasing the population size is similar to that of decreasing the learning rate.

Here, we organize the relation between the population size and learning rate more formally.
First, we examine the relation based on the results of quality gain analysis.
For the infinite-dimensional Sphere function, the optimal value of the normalized step-size $\bar{\sigma}^*$, whose normalized step-size is defined as $\bar{\sigma} := \sigma \eta_m d / \| m - x^* \| = \mathcal{O}(\sigma \eta_m)$,
is $\bar{\sigma}^* = - \mu_w \sum_{i=1}^{\lambda} w_i \E[\mathcal{N}_{i:\lambda}] \approx \sqrt{2 / \pi} \lambda \in \mathcal{O}(\lambda)$~\cite{opt_wr,qualitygainwres}.
Hence, the optimal step-size is $\sigma^{*} \in \mathcal{O}(\lambda / \eta_m)$, which clearly demonstrates that increasing $\lambda$ corresponds to decreasing $\eta_m$.
In other words, as the population size increases or learning rate decreases, the optimal step-size increases. \citet{cmasmallcm} hypothesized that CMA-ES with small learning rates can solve multimodal problems owing to the effect of maintaining a large step-size.

Next, we offer another characterization of the relation between the population size and the learning rate, by viewing IGO algorithms as discretizations of stochastic differential equations (SDEs)~\cite{jastrzkebski2017three}.
For conciseness, we define $g(\theta) := \E_{x \sim p(x;\theta)}[W_{\theta}^f (x) \tilde{\nabla}_{\theta}\ln p(x;\theta) ]$ and let its Monte Carlo estimation $\hat{g}^{(\lambda)}(\theta) := (1/\lambda) \sum_{i=1}^{\lambda} \hat{g}_i(\theta)$, where $\hat{g}_i(\theta) := W_{\theta}^f (x_i) \tilde{\nabla}_{\theta} \ln p (x_i; \theta)$, where $\hat{g}_i(\theta)$ is an unbiased estimator of $g(\theta)$.
Note that, in practice, $W_{\theta}^f$ must also be estimated using the Monte Carlo method;
thus, $\hat{g}_i(\theta)$ does not necessarily provide an unbiased estimation of $g(\theta)$.
However, we assume the availability of $W_{\theta}^f$ for this discussion.
Subsequently, we denote the covariance of $\hat{g}_i(\theta)$ as $S(\theta)$.
By using this notation, the IGO update in Eq.(\ref{eq:igo_update}) can be written as $\theta^{(t+1)} = \theta^{(t)} + \eta  \hat{g}^{(\lambda)}(\theta^{(t)})$.
Given a sufficiently large population size $\lambda$, the following is valid according to the central limit theorem:
\begin{align}
    \hat{g}^{(\lambda)}(\theta) \sim \mathcal{N}\left( g(\theta), \frac{1}{\lambda} S(\theta) \right).
\end{align}
Based on this result, we can rewrite Eq.(\ref{eq:igo_update}) as follows:
\begin{align}
    \theta^{(t+1)} = \theta^{(t)} + \eta g(\theta^{(t)}) + \eta (\hat{g}^{(\lambda)}(\theta^{(t)}) - g(\theta^{(t)})),
\end{align}
where $\hat{g}^{(\lambda)}(\theta^{(t)}) - g(\theta^{(t)}) \sim \mathcal{N}(0, (1/\lambda) S(\theta))$.
Hence, using the newly introduced random variable $\epsilon_{\theta} \sim \mathcal{N}(0, S(\theta))$, the IGO update can be rewritten as follows:
\begin{align}
    \label{eq:igo_eps}
    \theta^{(t+1)} = \theta^{(t)} + \eta g(\theta^{(t)}) + \frac{\eta}{\sqrt{\lambda}} \epsilon_{\theta^{(t)}}. 
\end{align}
Consequently, we consider the following SDE:
\begin{align}
    \label{eq:sde}
    d\theta = g(\theta) dt + \sqrt{\frac{\eta}{\lambda}} R(\theta) dW(t),
\end{align}
where $R(\theta) R(\theta)^{\top} = S(\theta)$ and $\{ W(t) \}$ is the standard Wiener process.
By discretizing the SDE using the Euler--Maruyama method~\cite{kloeden1992stochastic}, with the learning rate $\eta$, we obtain an equation identical to Eq.~(\ref{eq:igo_eps}).
Therefore, from the SDE perspective, the learning rate and the population size appear only in the form of the ratio $\eta / \lambda$, which implies that the effect of increasing $\lambda$ is similar to that of decreasing $\eta$.

In summary, although previous studies primarily adjusted the population size for solving multimodal problems, we empirically and theoretically observed that increasing the population size and decreasing the learning rate have similar effects on the optimal step-size and noise.

\subsection{Effect of Decreasing the Learning Rate from an ODE Perspective}
\label{sec:decreasing_lr}

\begin{figure*}[tb]
  \centering
  \includegraphics[width=0.564\hsize]{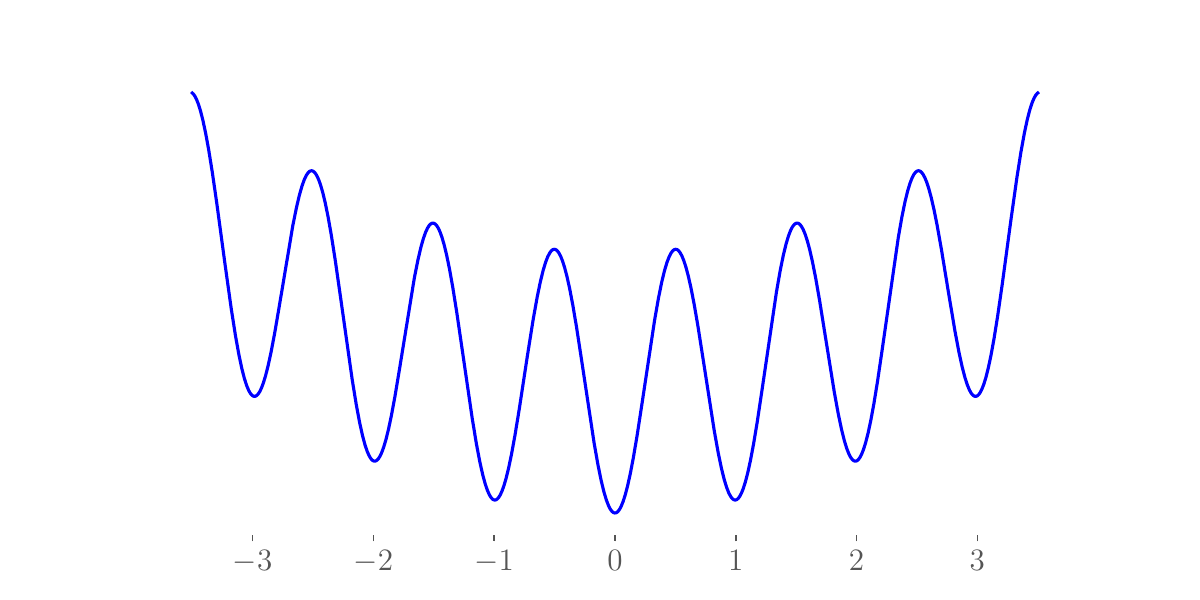}
  \caption{Rastrigin function.}
  \label{fig:rastrigin}
\end{figure*}

When the learning rate approaches zero, the IGO algorithm is reduced to the following ODE~\cite{akimoto2022ode}:
\begin{align}
    \frac{d\theta}{dt} &= \mathbb{E}_{x \sim p(x;\theta)}[W_{\theta}^f(x) \tilde{\nabla}_{\theta} \ln p(x;\theta) ] .
\end{align}
To illustrate the algorithm behavior from an ODE perspective, we consider minimizing the $1$-dimensional Rastrigin function $f_{\rm Rastrigin}(x) = 10 + x^2 - 10 \cos(2\pi x)$, which is a well-structured multimodal problem (Fig.~\ref{fig:rastrigin}).
Assuming that $W_{\theta}^f = -f$ and parameterizing our Gaussian distribution using $\theta = (m, v)$, where $m$ is the mean and $v$ is the variance, the ODEs are calculated as follows:
\begin{align}
    \label{eq:dmdt}
    \frac{dm}{dt} &= -2mv - 20\pi v \sin(2\pi m) \exp(-2\pi^2 v), \\
    \label{eq:dvdt}
    \frac{dv}{dt} &= -2v^2 - 40 \pi^2 v^2 \cos(2\pi m) \exp(-2\pi^2 v).
\end{align}

\begin{figure}[t]
  \centering
  \begin{tabular}{cc}
  \begin{minipage}{0.48\hsize}
  \centering%
  \includegraphics[width=\hsize]{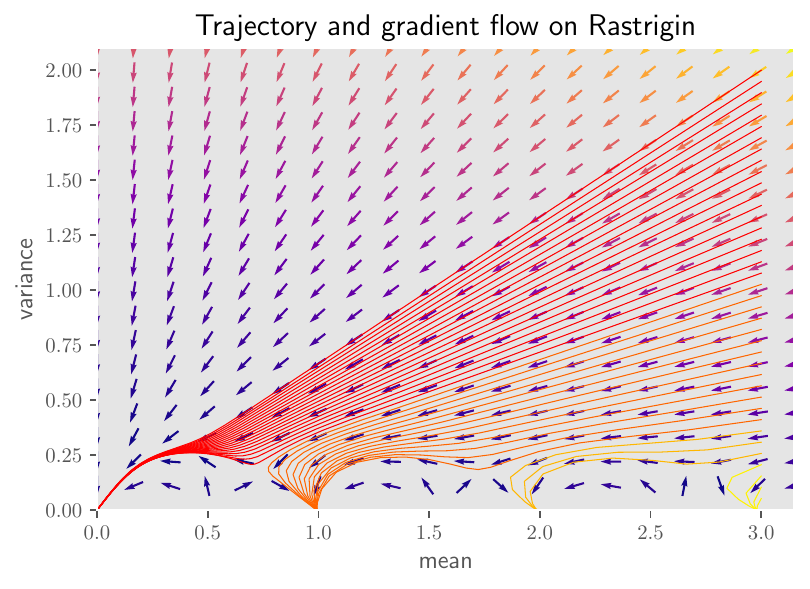}%
  \caption{ODE trajectories and gradient flows of the Rastrigin function. The different colors (\textcolor{red}{red}, \textcolor{orange}{orange}, \textcolor{yelloworange}{yellow-orange}, and \textcolor{yellow}{yellow}) of the ODE trajectories indicate different attractors.}%
  \label{fig:traj_gradflow}%
  \end{minipage}%
  &
  \begin{minipage}{0.48\hsize}%
  \centering%
  \includegraphics[width=\hsize]{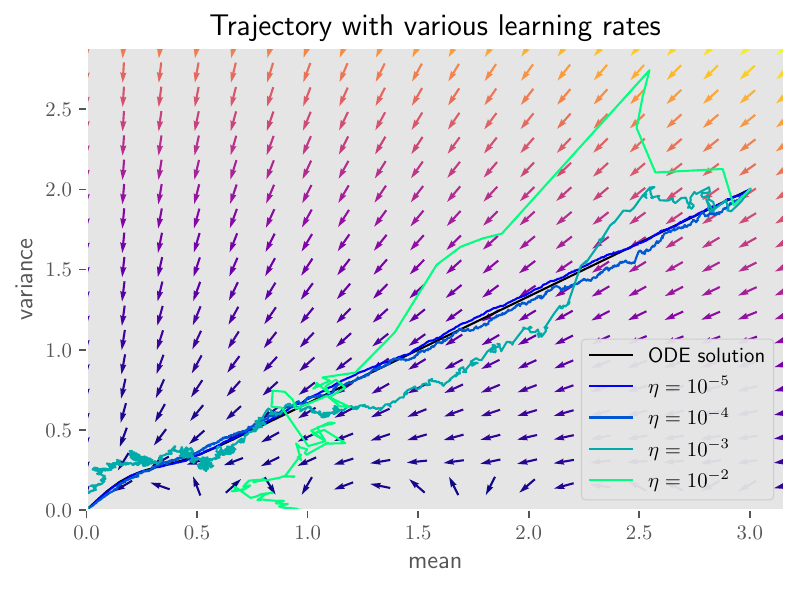}%
  \caption{Typical parameter trajectories of the Rastrigin function under various learning rates ($\eta = 10^{-5}, 10^{-4}, 10^{-3}, 10^{-2}$). The ODE solution (black) is also illustrated for reference.
  }%
  \label{fig:various_eta}%
  \end{minipage}%
  \end{tabular}
\end{figure}

\Cref{fig:traj_gradflow} shows the ODE trajectories and gradient flows of the Rastrigin function.
The experiments were conducted using initial distribution parameters $m=3.0$ and $v \in [0.02, 2.0]$.
It is evident that ODEs with large initial variances exhibit trajectories converging to the optimal solution $(m^*, v^*) = (0, 0)$.
Given that the algorithm behavior tends to approach the trajectory of the corresponding ODE, as the learning rate decreases, we hypothesize that such multimodal problems can be solved by adequately decreasing the learning rate and employing a sufficiently large variance.

To verify this hypothesis, we evaluated the behavior of the distribution parameters for various learning rates.
For this, we employed the following discretized versions of Eq. ~(\ref{eq:dmdt}) and (\ref{eq:dvdt}) using the Euler method:
\begin{align}
    m^{(t+1)} &= m^{(t)} - \eta (2mv + 20\pi v \sin(2\pi m) \exp(-2\pi^2 v)), \\
    v^{(t+1)} &= v^{(t)} - \eta (2v^2 + 40 \pi^2 v^2 \cos(2\pi m) \exp(-2\pi^2 v)),
\end{align}
where $\eta$ denotes the learning rate;
we used $\eta$ values of $10^{-5}, 10^{-4}, 10^{-3}$, and $10^{-2}$.
The initial distribution parameters were set as $(m, v) = (3.0, 2.0)$.
Figure~\ref{fig:various_eta} shows the typical behaviors of the parameter trajectories for various learning rates.
It is evident that as the learning rate decreases, the corresponding trajectory approaches the ODE solution, which is also evident from the design of the Euler method.
However, as the learning rate increases, the trajectory deviates from the ODE trajectory and tends to become trapped in the local optima, failing to find the optimal solution.
These findings suggest the importance of setting a small learning rate for multimodal problems that can be solved by moving the distribution parameters along the ODE trajectory.

Although earlier discussions focused on multimodal problems, we believe that decreasing the learning rate is equally important for problems with unbiased additive noise, represented as $f(x) + \epsilon$, where $\epsilon$ is an unbiased random variable, that is, $\mathbb{E}[\epsilon] = 0$.
This is because, in cases with unbiased noise, the corresponding ODE remains unchanged compared with noiseless ones.
That is, by decreasing the learning rate and aligning the parameter updates with the corresponding ODE trajectory, the distribution-parameter value can be guided closer to the optimal solution.

\subsection{Optimal Learning Rate}
\label{sec:optimal_lr}
Although setting a small learning rate can be beneficial for solving multimodal and noisy problems, as discussed in Section~\ref{sec:decreasing_lr}, using an excessively small value can result in slow convergence.
In this section, we explore the optimal value of the learning rate.

For simplicity, we consider the minimization of $\E[f(x)] = \int f(x) p(x;\theta) dx =: J(\theta)$ and assume that $J$ is twice differentiable.
Additionally, we let $\Delta$ be an unbiased estimator of $\tilde{\nabla} J(\theta)$.
In this case, the one-step update is $\theta - \eta \cdot \Delta$.
Using the Taylor approximation, we obtain the following:
\begin{align}
    J(\theta - \eta \cdot \Delta) &= J(\theta) - \eta \nabla J(\theta)^{\top} \Delta + \frac12 \eta^2 \Delta^{\top} H \Delta + o(\eta^2 \| \Delta \|^2) \\
    &\approx J(\theta) - \eta \nabla J(\theta)^{\top} \Delta + \frac12 \eta^2 \Delta^{\top} H \Delta,
\end{align}
where $H := \nabla^2 J(\theta)$.
Considering the expectations over $\Delta$, we obtain the following:
\begin{align}
    \E_{\Delta} [J(\theta - \eta \cdot \Delta)] \approx J(\theta) - \eta \nabla J(\theta)^{\top} \tilde{\nabla}J(\theta) + \frac12 \eta^2 \left( \tilde{\nabla}J(\theta)^{\top} H \tilde{\nabla}J(\theta) + \Tr(H \Cov[\Delta]) \right).
\end{align}
Thereafter, the expected improvement is approximated as follows:
\begin{align}
   J(\theta) -  \E_{\Delta} [J(\theta - \eta \cdot \Delta)] \approx \eta \nabla J(\theta)^{\top} \tilde{\nabla}J(\theta) - \frac12 \eta^2 \left( \tilde{\nabla}J(\theta)^{\top} H \tilde{\nabla}J(\theta) + \Tr(H \Cov[\Delta]) \right).
\end{align}
By taking the derivative w.r.t. $\eta$ and solving it for zero, we approximate the \emph{optimal} learning rate as follows:
\begin{align}
    \nabla J(\theta)^{\top}& \tilde{\nabla}J(\theta) - \eta \left( \tilde{\nabla}J(\theta)^{\top} H \tilde{\nabla}J(\theta) + \Tr(H \Cov[\Delta]) \right) = 0 \\
   &\therefore \eta^{*} \approx \frac{\nabla J(\theta)^{\top} \tilde{\nabla}J(\theta)}{\tilde{\nabla}J(\theta)^{\top} H \tilde{\nabla}J(\theta) + \Tr(H \Cov[\Delta])} \\
   &\qquad = \frac{\| \tilde{\nabla}J(\theta) \|_F^2}{\| \tilde{\nabla}J(\theta) \|_{H}^2 + \Tr(H \Cov[\Delta])},
\end{align}
where $F$ is the Fisher information matrix of the $\theta$, and $\norm{ \tilde{\nabla}J(\theta) }_M = (\tilde{\nabla}J(\theta)^\T M \tilde{\nabla}J(\theta))^{1/2}$ is the norm under $M$.

To obtain crucial insights into the determination of the optimal learning rate, we first assume $H \approx c F$ for a positive constant $c$.
This assumption is partially relevant in scenarios where the covariance matrix of CMA-ES successfully learns the shape of a quadratic function.
This concept can be illustrated as follows:
For a function $f(x) = \frac12 x^{\top}Ax$, the Hessian $H$ is $\diag(A, 0)$.
Consequently, given that $F = \diag(\Sigma^{-1}, \Sigma^{-1} \otimes \Sigma^{-1} / 2)$, if $A \propto \Sigma^{-1}$, then, to a certain extent, the Hessian $H$ in the $m$-part approximates $c F$ for some $c$ value; however, this does not apply to $H$ in the $\Sigma$-part.
Based on this assumption, the optimal learning rate can be written as 
\begin{align}
     \eta^{*} \approx \frac{1}{c} \cdot \frac{1}{1 + \mathrm{SNR}^{-1}} \propto \frac{1}{1 + \mathrm{SNR}^{-1}}.
\end{align}
where $\mathrm{SNR} := \frac{\| \tilde{\nabla}J(\theta) \|_F^2}{\Tr(F \Cov[\Delta])}$.
A high SNR increases the $\eta^{*}$ value, which aligns with our intuitive expectations.
In the next section, we propose an LRA mechanism based on these insights into the optimal learning rate.

\section{LRA Mechanism}
\label{sec:lra}

We consider the updating of the distribution parameters $\theta_m = m$ and $\theta_{\Sigma} = \vect (\Sigma)$, where $\vect$ is the vectorization operator and $\Sigma = \sigma^2 C$ for the standard CMA-ES.
Let $\Delta_{m}^{(t)} = m^{(t+1)} - m^{(t)}$ and $\Delta_{\Sigma}^{(t)} = \vect (\Sigma^{(t+1)} - \Sigma^{(t)})$ be the original updates of $m$ and $\Sigma$, respectively.
Subsequently, we introduce the learning rate factors $\eta_m^{(t)}$ and $\eta_\Sigma^{(t)}$.
The modified updates are performed as $\theta_m^{(t+1)} = \theta_m^{(t)} + \eta_m^{(t)} \Delta_m^{(t)}$ and $\theta_\Sigma^{(t+1)} = \theta_\Sigma^{(t)} + \eta_\Sigma^{(t)} \Delta_\Sigma^{(t)}$.
Finally, $\etam^{(t)}$ and $\etasig^{(t)}$ are adapted individually.

\subsection{Main Concept}

We adapt the learning rate factor $\eta$ for the component $\theta$ (either $\theta_m = m$ or $\theta_{\Sigma} = \vect(\Sigma)$) of the distribution parameters based on the SNR of the update as follows: 
\begin{equation}
    \mathrm{SNR} := \frac{\norm{\E[\Delta]}_{F}^2}{\Tr(F \Cov[\Delta])}
    = \frac{\norm{\E[\Delta]}_F^2}{\E[\norm{\Delta}^2_F] - \norm{\E[\Delta]}_F^2}.
\end{equation}
The Fisher metric is selected as it offers invariance against probability distribution parameterization.
We attempt to adapt $\eta$ such that $\mathrm{SNR} = \alpha \eta$, where $\alpha > 0$ is a hyperparameter that determines the target SNR.

The following rationale is employed for selecting this concept:
We assume that $\eta$ is sufficiently small such that the distribution parameters do not change significantly over $n$ iterations. 
Thus, we assume $\theta^{(t + k)} \approx \theta^{(t)}$ for $k = 1, \dots, n$. 
Subsequently, $\{\Delta^{(t + k)}\}_{k=0}^{n-1}$ are roughly considered as i.i.d. 
Hence, $n$ steps of the update are as follows:
\begin{subequations}
\begin{align}
\theta^{(t+n)} 
&= \theta^{(t)} + \eta \sum_{k=0}^{n-1} \Delta^{(t+k)}
\\
&\approx \theta^{(t)} + \mathcal{D}\left( n\eta \E[\Delta], n\eta^2 \Cov[\Delta] \right),
\end{align}
\end{subequations}
where $\mathcal{D}(A, B)$ is a distribution with expectation $A$ and (co)variance $B$.
Thus, by setting a small $\eta$ value and considering the results of $n = 1/\eta$ updates, we obtain an update that is more concentrated around the expected behavior than that expected for an update using $\eta = 1$.
The expected change in $\theta$ over $n = 1/\eta$ iterations, measured using the squared Fisher norm, which approximates the Kullback--Leibler (KL) divergence between $\theta^{(t)}$ and $\theta^{(t+n)}$, is $\norm{\E[\Delta]}_F^2 + \eta\Tr(F \Cov[\Delta])$, where the former and latter terms come from the signal and noise, respectively. 
The SNR over $n$ iterations is $\frac{\norm{\E[\Delta]}_F^2}{\eta\Tr(F \Cov[\Delta])} = \frac{1}{\eta} \mathrm{SNR}$.
Therefore, maintaining $\mathrm{SNR} = \alpha \eta$ implies maintaining the SNR at $\alpha$ over $n = 1/\eta$ iterations, independent of $\eta$. 

The rationale for using SNR can also be elucidated from the perspective of the optimal learning rate $\eta^{*}$  derived in Section~\ref{sec:optimal_lr}.
The results showed that $\eta^{*} \propto 1 / (1 + \mathrm{SNR}^{-1})$ approximately holds under some assumptions.
Additionally, we assume a relatively small SNR, for example, $\mathrm{SNR} \lessapprox 1$ (this assumption is validated in Appendix~\ref{sec:app_discuss_snr}).
In this case, the approximation $1 / (1 + \mathrm{SNR}^{-1}) \approx \mathrm{SNR}$ is roughly valid.
Thus, $\eta^{*} \propto \mathrm{SNR}$ can be considered to be valid.
As stated previously, we controlled $\eta$ such that $\mathrm{SNR} = \alpha \eta$.
Consequently, this leads to $\eta \propto \mathrm{SNR}$, which is considered to be nearly optimal.

\subsection{SNR Estimation}
\label{sec:snr_estimation}

We estimate $\norm{\E[\Delta]}^2$ and $\E[\norm{\Delta}^2]$ for each component ($m$ and $\Sigma$) using moving averages.
We let $\mvae^{(0)} = \bm{0}$ and $\mvav^{(0)} = 0$, and update them as follows:
\begin{subequations}
\begin{align}
    \label{eq:mvae_update}
    \mvae^{(t+1)} &= (1 - \beta) \mvae^{(t)} + \beta \tilde\Delta^{(t)} ,\\
    \label{eq:mvav_update}
    \mvav^{(t+1)} &= (1 - \beta) \mvav^{(t)} + \beta \norm{ \tilde\Delta^{(t)} }_2^2 , 
\end{align}
\end{subequations}
where $\beta$ is a hyperparameter; $\tilde\Delta^{(t)}$ is the update at iteration $t$ in the local coordinate at which the $F$ at $\theta^{(t)}$ becomes the identity; $\norm{\cdot}_2$ is the $\ell_2$-norm.
Thereafter, $\frac{2-\beta}{2 - 2\beta} \norm{\mvae}_2^2 - \frac{\beta}{2-2\beta} \mvav$ and $\mvav$ are considered estimates of $\norm{\E[\Delta]}_2^2$ and $\E[\norm{\Delta}_2^2]$, respectively (the derivation is included in Appendix~\ref{sec:app_snr}). 

The rationale for our estimators is as follows. 
Suppose that $\etam$ and $\etasig$ are sufficiently small for us to assume that the parameters $m$ and $\Sigma$ do not change significantly over $n$ iterations. 
Subsequently, the $\tilde\Delta^{(t+i)}\ (i=0,.., n-1)$ are considered to be located on the same local coordinates and distributed independently and identically. 
Then, ignoring the $(1 - \beta)^n$ terms, we obtain
\begin{align}
    \label{eq:mvae_dist}
    \mvae^{(t+n)} 
    \sim \mathcal{D}\left( \E[\tilde\Delta], \frac{\beta}{2 - \beta} \Cov[\tilde\Delta]\right).
\end{align}
(Again, the derivation is presented in Appendix~\ref{sec:app_snr}.)
Thus, we have $\E[\norm{\mvae}_2^2] \approx \norm{\E[\tilde\Delta]}_2^2 + \frac{\beta}{2-\beta}\Tr(\Cov[\tilde\Delta])$. 
Similarly, it is apparent that $\E[\mvav] \approx \E[\norm{\tilde\Delta}_2^2] = \norm{\E[\tilde\Delta]}_2^2 + \Tr(\Cov[\tilde\Delta])$.

The SNR is then estimated as
\begin{subequations}
\begin{align}
    \mathrm{SNR} &:= \frac{\norm{\E[\tilde\Delta]}^2}{\Tr(\Cov[\tilde\Delta])}
    = \frac{\norm{\E[\tilde\Delta]}^2}{\E[\norm{\tilde\Delta}^2] - \norm{\E[\tilde\Delta]}^2}
    \label{eq:snr_estimation_first}
    \\
    &\approx \frac{\norm{\mvae}_2^2 - \frac{\beta}{2-\beta} \mvav}{\mvav - \norm{\mvae}_2^2}
    =: \widehat{\mathrm{SNR}}.
    \label{eq:snr_estimation}
\end{align}
\end{subequations}

\subsection{Learning Rate Factor Adaptation}

We attempt to adapt $\eta$ such that $\widehat{\mathrm{SNR}} = \alpha \eta$, where $\alpha > 0$ is the hyperparameter. 
This adaptation is expressed as follows:
\begin{equation}
    \label{eq:eta}
    \eta \leftarrow \eta \exp\left( \min(\gamma\eta, \beta) \Pi_{[-1,1]}\left( \frac{\widehat{\mathrm{SNR}}}{\alpha \eta} - 1 \right)\right),
\end{equation}
where $\Pi_{[-1, 1]}$ is the projection onto $[-1, 1]$ and $\gamma$ is a hyperparameter.
If $\widehat{\mathrm{SNR}} > \alpha \eta$, $\eta$ increases, and vice versa.
Owing to these feedback mechanisms, $\widehat{\mathrm{SNR}} / (\alpha \eta)$ is expected to remain near $1$. 
In the above expression, the projection $\Pi_{[-1, 1]}$ is introduced to prevent a significant change in $\eta$ during an iteration, and the damping factor $\min(\gamma\eta, \beta)$ is introduced because of the following reasons.
First, the factor $\beta$ is introduced to allow for the effect of the change in the previous $\eta$ value to appear in $\widehat{\mathrm{SNR}}$. 
Second, the factor $\gamma\eta$ is introduced to prevent the $\eta$ value from changing more than $\exp(\gamma)$ or $\exp(-\gamma)$ over $1/\eta$ iterations. 
Based on the $\eta$ update through Eq.~(\ref{eq:eta}), the upper bound is set to $1$ using $\eta \leftarrow \min(\eta, 1)$, to prevent unstable behavior.
Although allowing $\eta$ values >$1$ would accelerate the optimization, we do not consider this because we aim to safely solve difficult problems.

\subsection{Local Coordinate-System Definition}
\label{sec:local_coordinate}

Although we estimate the SNR based on the updates $\Delta^{(\cdot)}$, na\"ively accumulating these updates $\Delta^{(\cdot)}$ may result in unintentional behavior, as illustrated in the following example.
Consider a scenario wherein $p(x;\theta^{(t)}) = \mathcal{N}(0, 100 I)$, $p(x;\theta^{(t+1)}) = \mathcal{N}(0, 50 I)$, and $p(x;\theta^{(t+2)}) = \mathcal{N}(0, 25 I)$.
In this case, the covariance matrix of the distribution decreases at a constant rate.
Consequently, each KL divergence is
$D_{\rm KL}(p(x;\theta^{(t)}) || p(x;\theta^{(t+1)})) = D_{\rm KL}(p(x;\theta^{(t+1)}) || p(x;\theta^{(t+2)}))$.
This implies that the distribution is moving at a uniform pace in terms of the KL divergence.
However, the updates are $\vect^{-1} (\Delta^{(t)}_{\Sigma}) = 50 I$ and $\vect^{-1} (\Delta^{(t+1)}_{\Sigma}) = 25 I$, whose scales are different.
Thus, accumulating these effects will result in unintentional behavior.

To address these issues, we ensure parameterization invariance by defining the local coordinate system~\cite{psaigo,psacmaes} such that the Fisher information matrices, $F_m$ and $F_{\Sigma}$, corresponding to each component of the distribution parameters, $m$ and $\Sigma$, respectively, are the identity matrices.
It is well-known that $F_m = \Sigma^{-1}$ and $F_{\Sigma} = 2^{-1} \Sigma^{-1} \otimes \Sigma^{-1}$, and their square roots are $\sqrt{F_m} = \sqrt{\Sigma}^{-1}$ and $\sqrt{F_{\Sigma}} = 2^{-\frac{1}{2}} \sqrt{\Sigma}^{-1} \otimes \sqrt{\Sigma}^{-1}$.
Therefore, we define
\begin{subequations}
\begin{align}
    \tilde{\Delta}_m &= \sqrt{\Sigma}^{-1} \Delta_m, \\
    \tilde{\Delta}_{\Sigma} &= 2^{-\frac{1}{2}} \vect(\sqrt{\Sigma}^{-1} \vect^{-1}(\Delta_{\Sigma}) \sqrt{\Sigma}^{-1}).
\end{align}
\end{subequations}
Actually, in the previous example, the local coordinate system allows us to easily verify that
$\vect^{-1} (\tilde{\Delta}^{(t)}_{\Sigma}) = \vect^{-1}(\tilde{\Delta}^{(t+1)}_{\Sigma})$.
This observation aligns with intuitive expectations in view of the KL divergence and suggests the validity of accumulating the updates $\tilde{\Delta}$ instead of the original $\Delta$.

\subsection{Covariance Matrix Decomposition}
After updating the covariance matrix $\Sigma^{(t+1)} = \Sigma^{(t)} + \eta_{\Sigma}^{(t)} \vect^{-1} (\Delta_{\Sigma}^{(t)})$, it must be split into $\sigma$ and $C$. 
For this, we adopt the following strategy:
\begin{subequations}
\begin{align}
    \sigma^{(t+1)} &= \det(\Sigma^{(t+1)})^{\frac{1}{2d}}, \\
    C^{(t+1)} &= (\sigma^{(t+1)})^{-2} \Sigma^{(t+1)}.   
\end{align}
\end{subequations}

\subsection{Step-size Correction}

Updating the learning rate for the $m$, i.e., $\eta_m$, changes the appropriate $\sigma$.
Through a quality gain analysis that analyzed the expected $f$ value improvement in a single step, a previous study~\cite{qualitygainwres} demonstrated that the optimal $\sigma$ value is proportional to $1/\eta_m$ for infinite-dimensional convex quadratic functions.
Therefore, to maintain the optimal $\sigma$ value under $\eta_m$ variations, we correct $\sigma$ after each $\eta_m$ update as follows:
\begin{equation}
\sigma^{(t+1)} \leftarrow \frac{\eta_m^{(t)}}{\eta_{m}^{(t+1)}} \sigma^{(t+1)} .
\end{equation}

\subsection{Overall Procedure}

Algorithm~\ref{alg:cma_learningrate} presents the overall LRA-CMA-ES procedure.
At Line 2, the old parameters $m^{(t)}, \sigma^{(t)},$ and $C^{(t)}$ are input into {\sf{CMA($\cdot$)}}, which outputs new parameters $m^{(t+1)}, \sigma^{(t+1)},$ and $C^{(t+1)}$ by executing Steps 1--3 described in Section~\ref{sec:cma}.

Note that the internal parameters such as the evolution paths $p_{\sigma}$ and $p_c$, are updated and stored in {\sf{CMA($\cdot$)}}. However, these values were omitted for simplicity.
The subscript $\cdot_{\{ m,\Sigma \}}$ (e.g., as in $\eta_{\{ m,\Sigma \}}$) indicates that there are parameters for $m$ and $\Sigma$, respectively.
For example, $\mvae^{(t+1)}_{\{ m,\Sigma \}} \gets (1 - \beta_{\{ m,\Sigma \}}) \mvae^{(t)}_{\{ m,\Sigma \}} + \beta_{\{ m,\Sigma \}} \tilde\Delta^{(t)}_{\{ m,\Sigma \}}$ is an abbreviation for the following two update equations: $\mvae^{(t+1)}_{m} \gets (1 - \beta_{m}) \mvae^{(t)}_{m} + \beta_{m} \tilde\Delta^{(t)}_{m}$ and $\mvae^{(t+1)}_{\Sigma} \gets (1 - \beta_{\Sigma}) \mvae^{(t)}_{\Sigma} + \beta_{\Sigma} \tilde\Delta^{(t)}_{\Sigma}$.

\begin{algorithm}
\caption{LRA-CMA-ES}
\label{alg:cma_learningrate}
\begin{algorithmic}[1]
\Require $m^{(0)} \in \mathbb{R}^d, \sigma^{(0)} \in \mathbb{R}_{>0}, \lambda \in \mathbb{N}, \alpha, \beta_{\{ m,\Sigma \}}, \gamma \in \mathbb{R}$
\Set $t = 0, C^{(0)}=I, \eta_{\{m,\Sigma\}}^{(0)} = 1, \mvae^{(0)} = \bm{0}, \mvav^{(0)} = 0$
\While{stopping criterion not met}
    \State $m^{(t+1)}, \sigma^{(t+1)}, C^{(t+1)} \gets {\rm CMA}(m^{(t)}, \sigma^{(t)}, C^{(t)})$
    \State {\sf \//\// calculate parameter one-step differences}
    \State $\Delta_m^{(t)} \gets m^{(t+1)} - m^{(t)}$
    \State $\Sigma^{(t+1)} \gets \left( \sigma^{(t+1)} \right)^2 C^{(t+1)}$
    \State $\Delta_{\Sigma}^{(t)} \gets {\rm vec}\left( \Sigma^{(t+1)} - \Sigma^{(t)} \right)$
    \State {\sf \//\// local coordinate}
    \State $\tilde{\Delta}_m^{(t)} \gets \sqrt{\Sigma^{(t)}}^{-1} \Delta_m^{(t)}$
    \State $\tilde{\Delta}_{\Sigma}^{(t)} \gets 2^{-1/2} {\rm vec}\left(\sqrt{\Sigma^{(t)}}^{-1} {\rm vec}^{-1}\left(\Delta_{\Sigma}^{(t)}\right) \sqrt{\Sigma^{(t)}}^{-1} \right)$
    \State {\sf \//\// update evolution paths and estimate SNR}
    \State $\mvae^{(t+1)}_{\{ m,\Sigma \}} \gets (1 - \beta_{\{ m,\Sigma \}}) \mvae^{(t)}_{\{ m,\Sigma \}} + \beta_{\{ m,\Sigma \}} \tilde\Delta^{(t)}_{\{ m,\Sigma \}}$
    \State $\mvav^{(t+1)}_{\{ m,\Sigma \}} \gets (1 - \beta_{\{ m,\Sigma \}}) \mvav^{(t)}_{\{ m,\Sigma \}} + \beta_{\{ m,\Sigma \}} \norm{ \tilde\Delta^{(t)}_{\{ m,\Sigma \}} }_2^2$
    \State $\widehat{\mathrm{SNR}}_{\{ m,\Sigma \}} \gets \frac{\norm{\mvae^{(t+1)}_{\{ m,\Sigma \}} }_2^2 - \frac{\beta_{\{ m,\Sigma \}} }{2-\beta_{\{ m,\Sigma \}} } \mvav^{(t+1)}_{\{ m,\Sigma \}} }{\mvav^{(t+1)}_{\{ m,\Sigma \}} - \norm{\mvae^{(t+1)}_{\{ m,\Sigma \}}}_2^2}$
    \State {\sf \//\// update learning rates}
    \State $\eta_{\{ m,\Sigma \}}^{(t+1)} \leftarrow \eta_{\{ m,\Sigma \}}^{(t)}$
    \Statex $\cdot \exp\left( \min(\gamma \eta_{\{ m,\Sigma \}}^{(t)}, \beta_{\{ m,\Sigma \}}) \Pi_{[-1,1]}\left( \frac{\widehat{\mathrm{SNR}}_{\{ m,\Sigma \}} }{\alpha \eta_{\{ m,\Sigma \}} } - 1 \right)\right)$
    \State $\eta_{\{ m,\Sigma \}}^{(t+1)} \leftarrow \min(\eta_{\{ m,\Sigma \}}^{(t+1)}, 1)$
    \State {\sf \//\// update parameters with adaptive learning rates}
    \State $m^{(t+1)} \gets m^{(t)} + \eta_m^{(t+1)} \Delta_m^{(t)}$
    \State $\Sigma^{(t+1)} \gets \Sigma^{(t)} + \eta_{\Sigma}^{(t+1)} \vect^{-1} (\Delta_{\Sigma}^{(t)})$
    \State {\sf \//\// decompose $\Sigma$ to $\sigma$ and $C$}
    \State $\sigma^{(t+1)} \gets \det(\Sigma^{(t+1)})^{\frac{1}{2d}}$, $C^{(t+1)} \gets (\sigma^{(t+1)})^{-2} \Sigma^{(t+1)}$
    \State {\sf \//\// $\sigma$ correction}
    \State $\sigma^{(t+1)} \gets \sigma^{(t+1)} (\eta_{m}^{(t)} / \eta_{m}^{(t+1)})$
    \State $t \gets t + 1$
\EndWhile
\end{algorithmic}
\end{algorithm}

\section{Experiments}
\label{sec:exp}
This study included various experiments to investigate the following research questions (RQs):

\begin{itemize}
    \item [\textbf{RQ1.}] Does the $\eta$ adaptation in LRA-CMA-ES behave appropriately in accordance with the problem structure?
    \item [\textbf{RQ2.}] Can LRA-CMA-ES solve multimodal and noisy problems even though a default $\lambda$ value is used? How does its efficiency compare to that of CMA-ES with a fixed $\eta$ value?
    \item [\textbf{RQ3.}] How does the performance change with changes in LRA-CMA-ES hyperparameters?
    \item [\textbf{RQ4.}] How does the performance depend on the population size $\lambda$?
    \item [\textbf{RQ5.}] What are the differences in the performances of LRA-CMA-ES, which adapts the learning rate, and PSA-CMA-ES~\cite{psacmaes}, which adapts the population size?
\end{itemize}

The remainder of this section is organized as follows.
The experimental setups are described in Section~\ref{sec:setup}.
Section~\ref{sec:lrbahavior} demonstrates $\eta$ adaptation in LRA-CMA-ES for noiseless and noisy problems (\textbf{RQ1}).
Section~\ref{sec:fixed_vs_adapt} compares LRA-CMA-ES with CMA-ES with fixed $\eta$ values (\textbf{RQ2}).
Section~\ref{sec:exp_hyperparam} investigates the effects of LRA-CMA-ES hyperparameters (\textbf{RQ3}).
Additional experimental results for the hyperparameters are presented in Appendix~\ref{sec:app_expresults}.
Section~\ref{sec:exp_popsize} evaluates the performance differences under various different population sizes (\textbf{RQ4}).
Finally, Section~\ref{sec:exp_vs_psa} compares LRA-CMA-ES with PSA-CMA-ES (\textbf{RQ5}).
Our code is available at \textcolor{blue}{https://github.com/nomuramasahir0/cma-learning-rate-adaptation}.
The LRA-CMA-ES is also available at \textcolor{blue}{https://github.com/CyberAgentAILab/cmaes}~\cite{nomura2024cmaes}, although the CMA-ES implementation there differs slightly from the one in our paper.

\subsection{Experimental Setups}
\label{sec:setup}
The benchmark problem definitions and initial distributions are presented in Table~\ref{tab:benchmark}.
In each case (except for the Rosenbrock function), the global optimal solution is at $x = 0$. 
However, for the Rosenbrock function, it is at $x = 1$.
As unimodal problems, we employ the Sphere, Ellipsoid, and Rosenbrock functions.
The reason for using unimodal problems is to ensure that the performance of LRA-CMA-ES does not degrade significantly compared to CMA-ES with the default learning rate.
The Ellipsoid function is an ill-conditioned problem, whereas the Rosenbrock function has dependencies between variables.
Although the Rosenbrock function has local minima, in our study, it can be regarded as an almost unimodal problem.
As well-structured multimodal problems, we employ the Ackley, Schaffer, Rastrigin, Bohachevsky, and Griewank functions.
These problems are composed of a quadratic convex function and oscillatory nonconvex function, which resembles the noise.
In the Ackley, Rastrigin, Bohachevsky, and Griewank functions, the oscillatory function is added to the convex function, whereas in the Schaffer function, the oscillatory function affects the convex function multiplicatively.
This causes the Schaffer function to have fine oscillations around the optimal solution.
Noteworthy, the optimization for the Griewank function gets easier as the dimension increases~\cite{hansen2004evaluating}.
Similar to \cite{hansen2004evaluating}, we imposed additional bounds
on the Ackley function.
For noisy problems, we considered an additive Gaussian noise $\epsilon \sim \mathcal{N}(0, \sigma_n^2)$ with $\sigma_n^2$ variance.
It is worth noting that the proposed method maintains the affine invariance of CMA-ES because LRA-CMA-ES does not rely on a specific parameterization, as discussed in Section~\ref{sec:local_coordinate}.
Consequently, although many of the employed benchmark problems are separable, the experimental results obtained from a benchmark function can still be generalized to the experimental results of its rotated version.

In all the experiments (except for those in Section~\ref{sec:exp_popsize}), we set the default $\lambda = 4 + \lfloor 3 \ln d \rfloor$.
The result when changing $\lambda$ can be found in Ref.~\cite{hansen2004evaluating} and Section~\ref{sec:exp_popsize}.
For the dimension $d$, we employed $d \in \{ 10, 20, 30, 40 \}$ for noiseless problems and $d = 10$ for noisy problems.
Additionally, we set the LRA-CMA-ES hyperparameters as $\alpha = 1.4$, $\beta_m = 0.1$, $\beta_{\Sigma} = 0.03$, and $\gamma = 0.1$ based on preliminary experiments.
As noted above, Section~\ref{sec:exp_hyperparam} presents an analysis of the hyperparameters sensitivity.
The values of other internal parameters of CMA-ES were set to those recommended in \cite{hansen2014principled}.

{\footnotesize
\begin{table*}[t]
  \centering
  \caption{Definitions of benchmark problems and initial distributions used in the experiments.}
  \label{tab:benchmark}
  \begin{tabular}{l|l}
    \bottomrule
    Definitions & Initial Distributions \\
    \hline \hline
    $f_{\rm Sphere}(x) = \sum_{i=1}^{d} x_i^2$ & $m^{(0)} = [3,\ldots,3], \sigma^{(0)} = 2$ \\
    $f_{\rm Ellipsoid}(x) = \sum_{i=1}^{d} (1000^{\frac{i-1}{d-1}}x_i)^2$ & $m^{(0)} = [3,\ldots,3], \sigma^{(0)} = 2$ \\
     $f_{\rm Rosenbrock}(x) = \sum_{i=1}^{d-1} (100 (x_{i+1} - x_i^2)^2 + (x_i - 1)^2)$ & $m^{(0)} = [0,\ldots,0], \sigma^{(0)} = 0.1$ \\
     $f_{\rm Ackley}(x) = 20 - 20 \cdot \exp (-0.2 \sqrt{\frac{1}{d} \sum_{i=1}^d x_i^2}) + e - \exp (\frac{1}{d} \sum_{i=1}^d \cos(2\pi x_i))$ & $m^{(0)} = [15.5,\ldots,15.5], \sigma^{(0)} = 14.5$ \\
     $f_{\rm Schaffer}(x) = \sum_{i=1}^{d-1} (x_i^2 + x_{i+1}^2)^{0.25} \cdot [\sin^2(50\cdot (x_i^2 + x_{i+1}^2)^{0.1}) + 1]$ & $m^{(0)} = [55,\ldots,55], \sigma^{(0)} = 45$ \\
     $f_{\rm Rastrigin}(x) = 10d + \sum_{i=1}^d (x_i^2 - 10 \cos(2\pi x_i))$ & $m^{(0)} = [3,\ldots,3], \sigma^{(0)} = 2$ \\
     $f_{\rm Bohachevsky}(x) = \sum_{i=1}^{d-1} ( x_i^2 + 2x_{i+1}^2 - 0.3 \cos(3\pi x_i) - 0.4\cos(4\pi x_{i+1}) + 0.7 )$ & $m^{(0)} = [8,\ldots,8], \sigma^{(0)} = 7$ \\
     $f_{\rm Griewank}(x) = \frac{1}{4000} \sum_{i=1}^d x_i^2 - \Pi_{i=1}^d \cos (x_i / \sqrt{i}) + 1$ & $m^{(0)} = [305,\ldots,305], \sigma^{(0)} = 295$ \\
    \toprule
  \end{tabular}
\end{table*}
}

\begin{figure*}[tb]
  \centering
  \includegraphics[width=0.964\hsize,trim=5 5 5 5,clip]{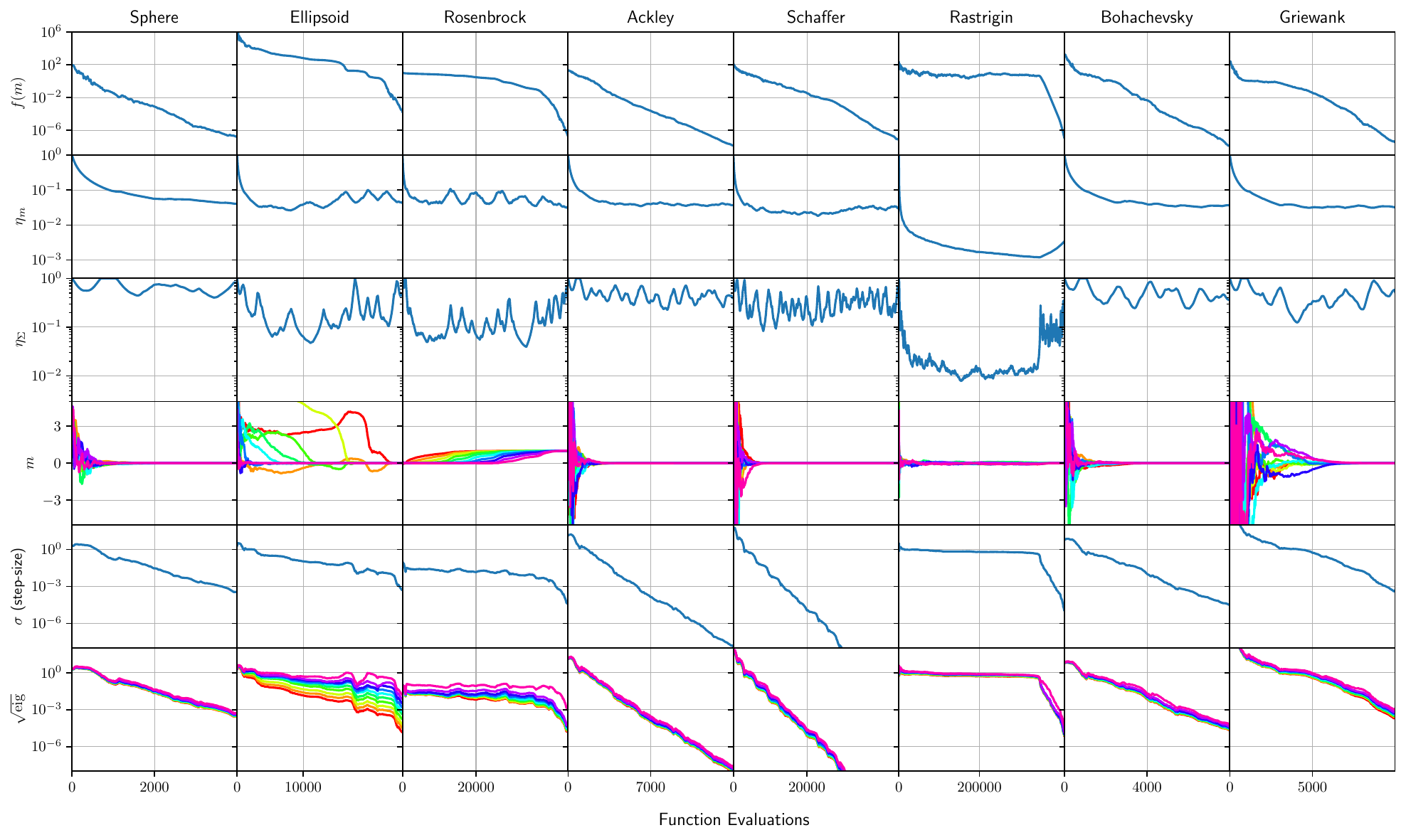}
  \caption{Typical LRA-CMA-ES behaviors for 10-dimensional (10-D) noiseless problems. The coordinates of $m$ and the square roots of the eigenvalues of $\sigma^2 C$ (denoted by $\sqrt{\text{eig}}$) are indicated with different colors.}
  \label{fig:adaptlr_behavior}
\end{figure*}

\begin{figure}[tb]
  \centering
  \includegraphics[width=0.564\hsize,trim=5 5 5 5,clip]{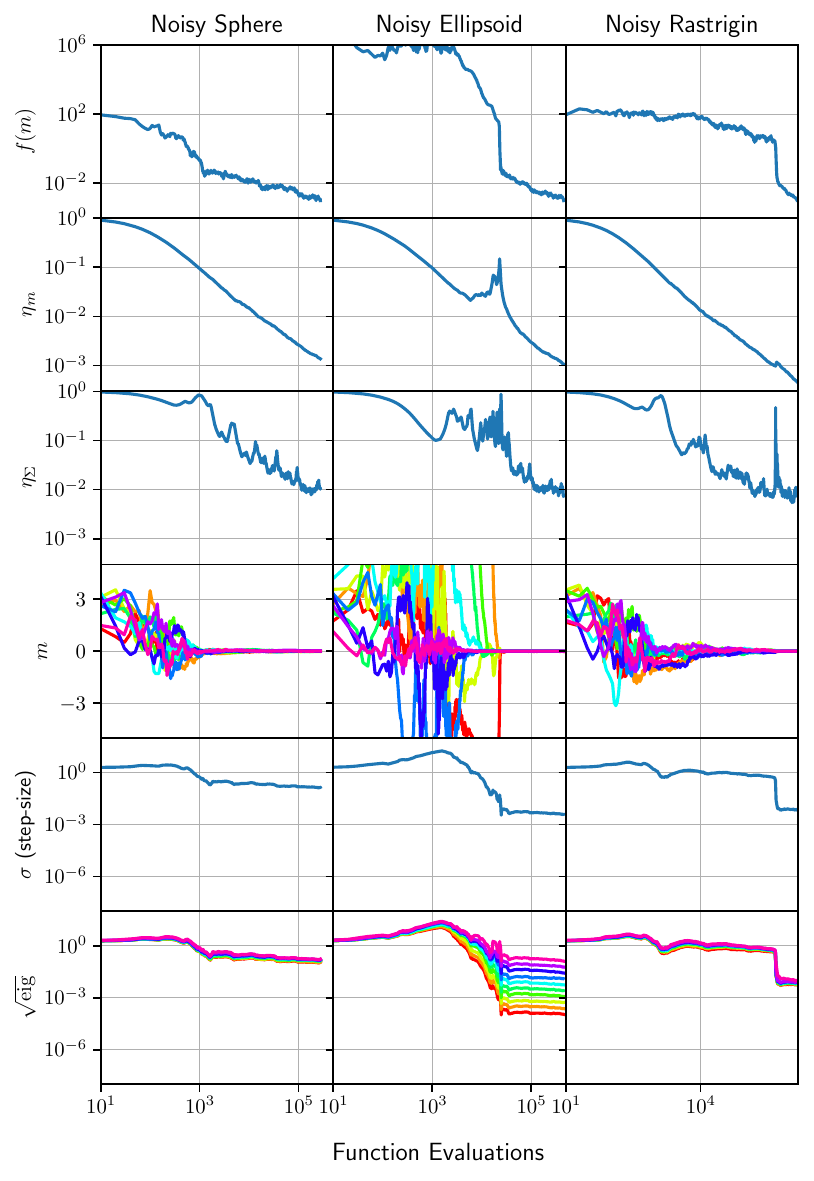}
  \caption{Typical LRA-CMA-ES behaviors for 10-D noisy problems. The noise variance $\sigma_n^2$ was set to $1$.}
  \label{fig:adaptlr_behavior_noisy_loglinear}
\end{figure}

\subsection{Learning Rate Behavior}
\label{sec:lrbahavior}
Figure~\ref{fig:adaptlr_behavior} shows the typical LRA-CMA-ES behaviors for noiseless problems, wherein $\eta_{\Sigma}$ maintained relatively large values for the Sphere function.
However, it exhibits significantly smaller values for the Ellipsoid and Rosenbrock functions.
We believe that this behavior is undesirable, because the default $\eta$ value already works well for these unimodal problems.
Although $\eta$ can be increased by changing the hyperparameters of the proposed $\eta$ adaptation, this change may be detrimental for multimodal problems.

It is evident that $\eta_m$ is slightly smaller for multimodal problems than for unimodal problems.
Particularly, for the Rastrigin function, $\eta_m$ and $\eta_{\Sigma}$ clearly decrease at the beginning of the optimization, which reflects the difficulty of multimodal problem optimization.
Subsequently, $\eta$ increases as optimization becomes as easy as that for a unimodal problem.
This behavior demonstrates that LRA-CMA-ES can adapt $\eta$ according to the search difficulty.

Figure~\ref{fig:adaptlr_behavior_noisy_loglinear} shows the typical $\eta$ adaptation behavior for noisy problems.
The noise has a negligible effect in the early stages; thus, the $\eta$ behavior for noisy problems is similar to that for noiseless problems.
However, as the optimization proceeds and the function value approaches the same scale as that of the noise value, the noise starts having a critical effect.
Consequently, the $\eta$ value decreases. This adaptation ensures that the SNR remains constant.
Notably, similar behavior can be observed for the noisy Rastrigin function, which features both noise and multimodality.

\subsection{Effects of LRA}
\label{sec:fixed_vs_adapt}

\begin{figure*}[tb]
  \centering
  \includegraphics[width=0.964\hsize,trim=5 5 5 5,clip]{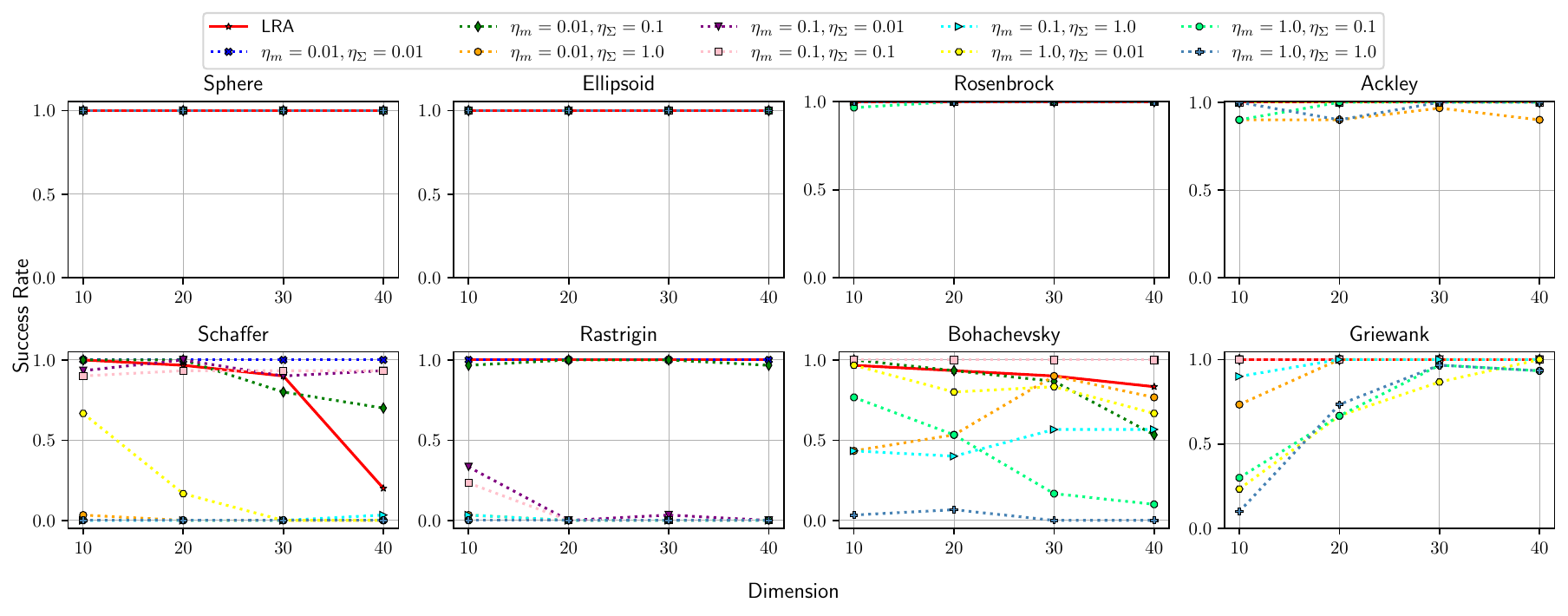}
  \caption{Success rates according to the number of dimensions (noiseless problems).}
  \label{fig:compare_succrate}
\end{figure*}

\begin{figure*}[tb]
  \centering
  \includegraphics[width=0.964\hsize,trim=5 5 5 5,clip]{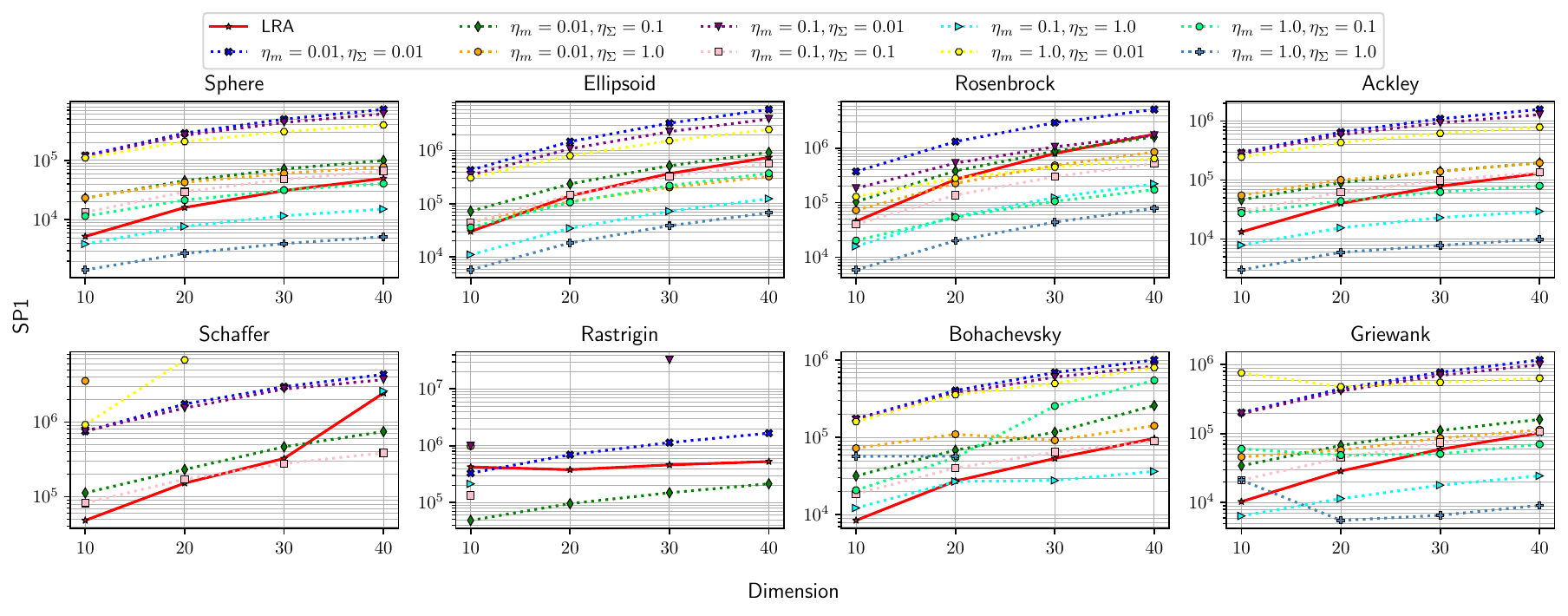}
  \caption{SP1 values according to the number of dimensions (noiseless problems). A missing point indicates the algorithm's failure in all trials.}
  \label{fig:compare_sp1}
\end{figure*}

Figures~\ref{fig:compare_succrate} and ~\ref{fig:compare_sp1} show the performances of LRA-CMA-ES and that of CMA-ES with a fixed learning rate $(\eta_m, \eta_{\Sigma} \in \{10^0, 10^{-1}, 10^{-2} \})$ for the noiseless problems.
Note that CMA-ES with $\eta_m = 1.0$ and $\eta_{\Sigma} = 1.0$ is the original CMA-ES with the default $\eta$ value.
Each trial was considered successful if $f(m)$ reached the target value $10^{-8}$ before $10^7$ function evaluations or before a numerical error occurred because of an excessively small $\sigma$.
In addition to the success rate, we employed the SP1 value~\cite{auger2005restart}, which is the average number of evaluations among successful trials until achieving the target value divided by the success rate.
30 trials were conducted for each setting.

To compare the performances of these strategies for the noisy problems, we employed the empirical cumulative density function (ECDF) of COCO, a platform for comparing continuous optimizers in a black-box setting~\cite{hansen2021coco}.
Using $N_{\rm target}$ target values, we recorded the number of evaluations until $f(m)$ (noiseless) reached each target value for the first time, and set the maximum function evaluation to $10^8$.
We collected data by running $N_{\rm trial}$ independent trials, and obtained a total of $N_{\rm target} \cdot N_{\rm trial}$ targets for each problem.
Thereafter, we set the target values to $10^{6 - 9(i-1)/(N_{\rm target} - 1)}$ for $i = 1, \ldots, N_{\rm target}$, with $N_{\rm target} = 30$. By executing $N_{\rm trial} = 20$ trials, 600 targets were obtained for each problem.
Figure~\ref{fig:ecdf} shows the target value percentages obtained for each number of evaluations.

\subsubsection{Noiseless Problems}
We compared the success rates of LRA-CMA-ES and CMA-ES with fixed $\eta$ values, as shown in Figure~\ref{fig:compare_succrate}.
For the multimodal problems, CMA-ES with a large $\eta$ often failed to reach the optimum.
However, CMA-ES with a small $\eta$ exhibited a high success rate, indicating a clear dependence on $\eta$.
By contrast, LRA-CMA-ES exhibited a relatively good success rate, even though no $\eta$ tuning was required.
It is noteworthy that LRA-CMA-ES succeeded in all trials for the Rastrigin function even though the default population size (e.g., $\lambda = 15$ for $d=40$) was used and $\eta$ was not tuned in advance.

However, LRA-CMA-ES performance for the Schaffer function degraded at $d=40$.
From the results indicating that CMA-ES with an appropriately tuned, small $\eta$ achieved a relatively high success rate, the LRA-CMA-ES result may have been obtained because $\eta$ was not appropriately adapted in that case.
This will be investigated in future work.

Figure~\ref{fig:compare_sp1} shows the SP1 results for LRA-CMA-ES and CMA-ES with fixed $\eta$ values.
CMA-ES with the default $\eta$ values ($\eta_m = 1.0, \eta_{\Sigma} = 1.0$) outperformed the other methods for unimodal problems; however, the performance degraded significantly for multimodal problems owing to optimization failures.
By contrast, the CMA-ES with a small $\eta$ sometimes exhibited good performance for such multimodal problems;
however, it was not efficient for unimodal and relatively easy multimodal problems.
Therefore, for CMA-ES with a fixed $\eta$ value, a clear trade-off in efficiency exists based on the $\eta$ setting.
By contrast, LRA-CMA-ES exhibited stable and relatively good performance for unimodal and multimodal problems.
Again, $\eta$ was not tuned, which is significantly expensive in practice.
There is scope for improvement of the LRA-CMA-ES performance on unimodal problems; however, the current sub-par performance can be somewhat mitigated by changing the hyperparameters.
The effects of the hyperparameters are discussed in Section~\ref{sec:exp_hyperparam}.

\subsubsection{Noisy Problems}

\begin{figure*}[tb]
  \centering
  \includegraphics[width=0.964\hsize,trim=5 5 5 5,clip]{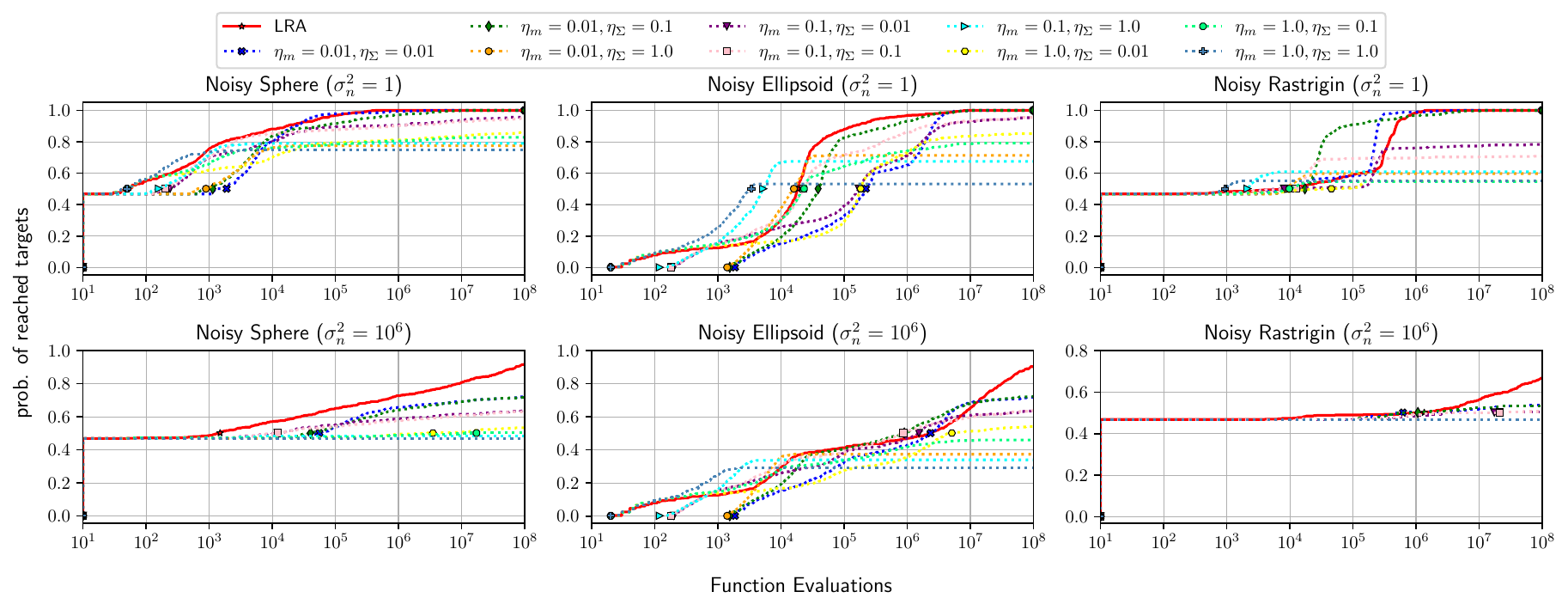}
  \caption{Empirical cumulative density function for 10-D noisy problems, with $\sigma_n^2$ set to $1$ or $10^6$.}
  \label{fig:ecdf}
\end{figure*}

Figure~\ref{fig:ecdf} shows the ECDF results for both LRA-CMA-ES and CMA-ES with fixed $\eta$ values.
We considered two noise strengths, weak and strong, that is, $\sigma_n^2 = 1$ and $10^6$, respectively.

Under the weak-noise setting, CMA-ES with a small $\eta$ value reached all the target values.
By contrast, CMA-ES with a large $\eta$ value failed to approach the global optimum and yielded a sub-optimal solution.
LRA-CMA-ES achieved similar performance to CMA-ES with a small $\eta$ value without tuning.
However, under the strong-noise setting, even CMA-ES with a small $\eta$ stopped improving the $f$ value before reaching the global optimum.
By contrast, LRA-CMA-ES continued improving the $f$ value.
Notably, the results for the noisy Rastrigin function suggest that LRA-CMA-ES can simultaneously handle both noise and multimodality.

\subsection{Effects of Hyperparameters}
\label{sec:exp_hyperparam}

\begin{figure*}[tb]
  \centering
  \includegraphics[width=0.964\hsize,trim=5 10 5 25,clip]{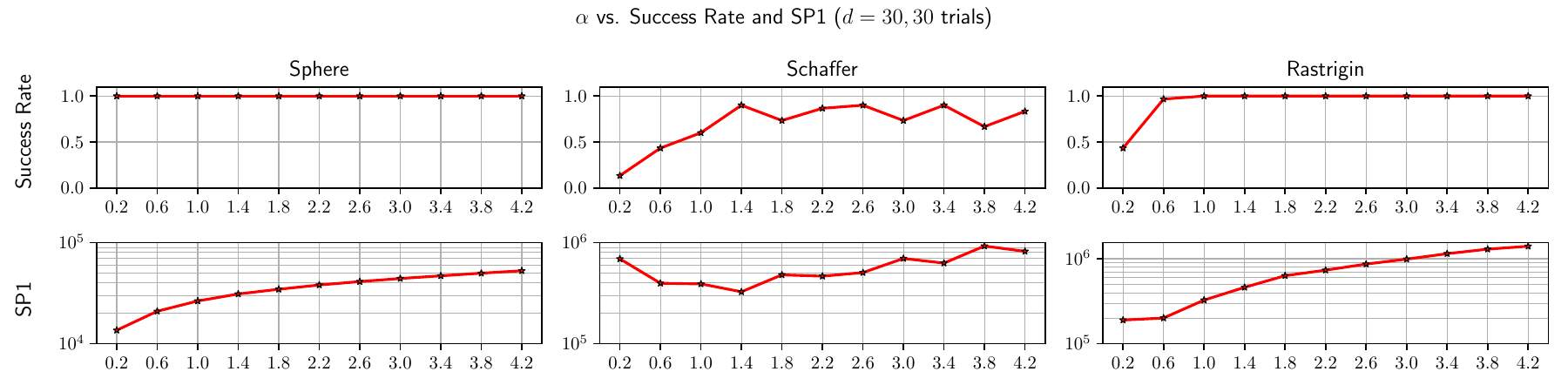}
  \caption{Success rates and SP1 values with hyperparameter $\alpha$ for 30-D noiseless problems (30 trials).}
  \label{fig:alpha_srsp1_d=30}
\end{figure*}

\begin{figure*}[tb]
  \centering
  \includegraphics[width=0.964\hsize,trim=5 10 5 25,clip]{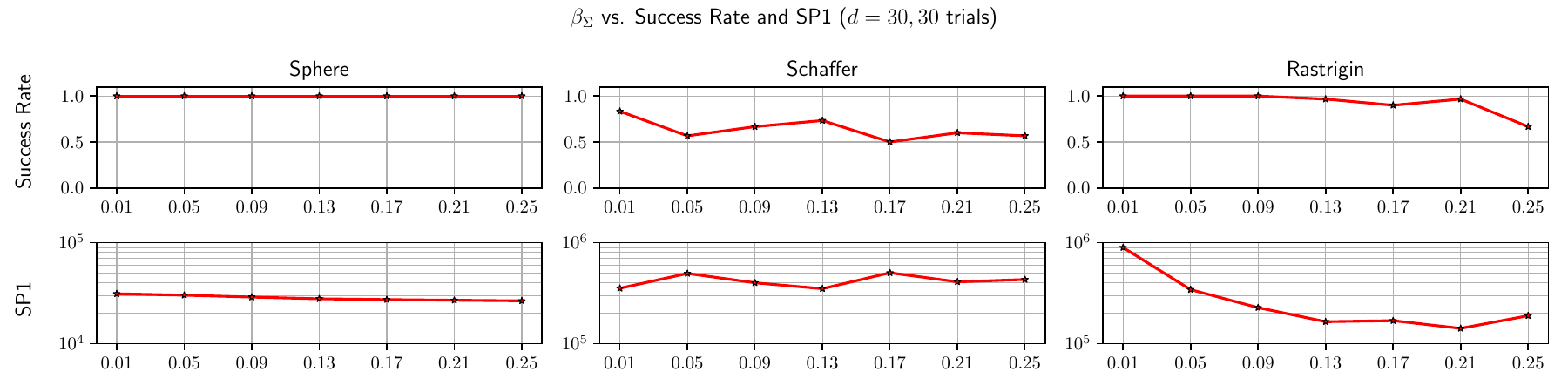}
  \caption{Success rates and SP1 values with hyperparameter $\beta_{\Sigma}$ for 30-D noiseless problems (30 trials).}
  \label{fig:beta_Sigma_srsp1_d=30}
\end{figure*}

Figure~\ref{fig:alpha_srsp1_d=30} shows the success rates and SP1 values with respect to $\alpha$ for the 30-dimensional (30-D) noiseless Sphere, Schaffer, and Rastrigin functions.
For the Sphere function, a lower SP1 value could be achieved with a smaller $\alpha$ value.
However, an excessively large $\alpha$ results in optimization failures for multimodal problems.
Therefore, the current setting of $\alpha = 1.4$ seems reasonable; however, further investigations are required.

Figure~\ref{fig:beta_Sigma_srsp1_d=30} shows the success rates and SP1 values with respect to $\beta_{\Sigma}$.
We observe that an excessively large $\beta_{\Sigma}$ causes optimization failures in the Rastrigin function.
Conversely, an excessively small $\beta_{\Sigma}$ results in slow convergence.
An additional result ($\beta_{\Sigma} \in \{ 0.01, 0.02, ..., 0.05 \}$) is presented in Appendix~\ref{sec:app_expresults}.

We also conducted similar experiments on the hyperparameters $\beta_m$ and $\gamma$, to confirm their effects.
These hyperparameters mildly impacted the overall performance compared to $\alpha$ and $\beta_{\Sigma}$
(these results are also presented in Appendix~\ref{sec:app_expresults}).

\subsection{Effects of Population Size}
\label{sec:exp_popsize}
Although we used the \emph{default} population size, $\lambda = 4 + \lfloor 3 \log (d) \rfloor$, in all the experiments, practitioners may want to employ different population sizes to fully utilize their parallel environments.
In this section, we describe the experiments conducted to investigate the effects of population size.

\begin{figure*}[tb]
  \centering
  \includegraphics[width=0.964\hsize,trim=5 10 5 25,clip]{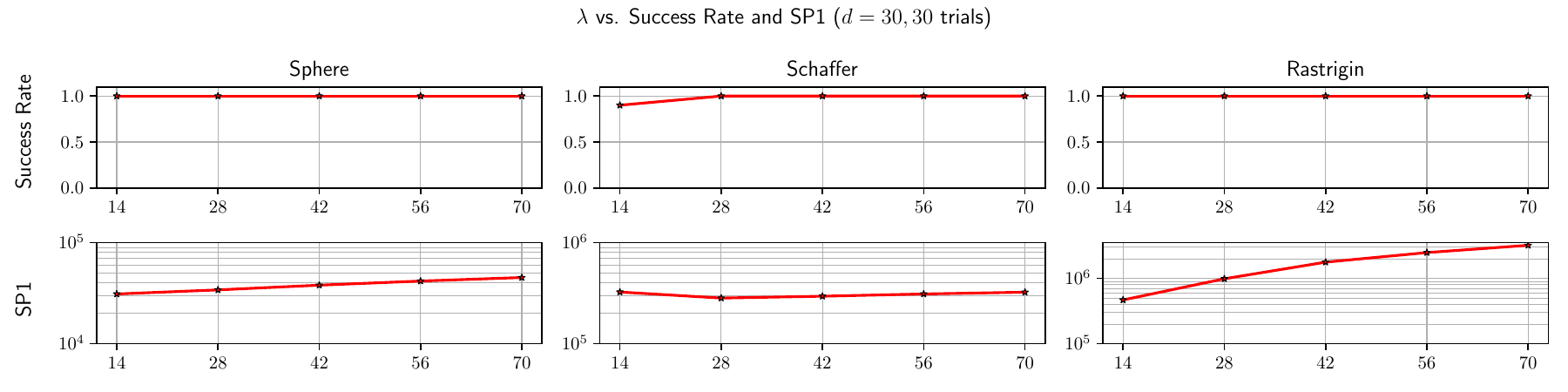}
  \caption{Success rates and SP1 values with $\lambda \in \{ 14, 28, 42, 56, 70 \}$ for 30-D noiseless problems (30 trials).}
  \label{fig:lamb_srsp1_d=30_small}
\end{figure*}

\begin{figure*}[tb]
  \centering
  \includegraphics[width=0.964\hsize,trim=5 0 5 0,clip]{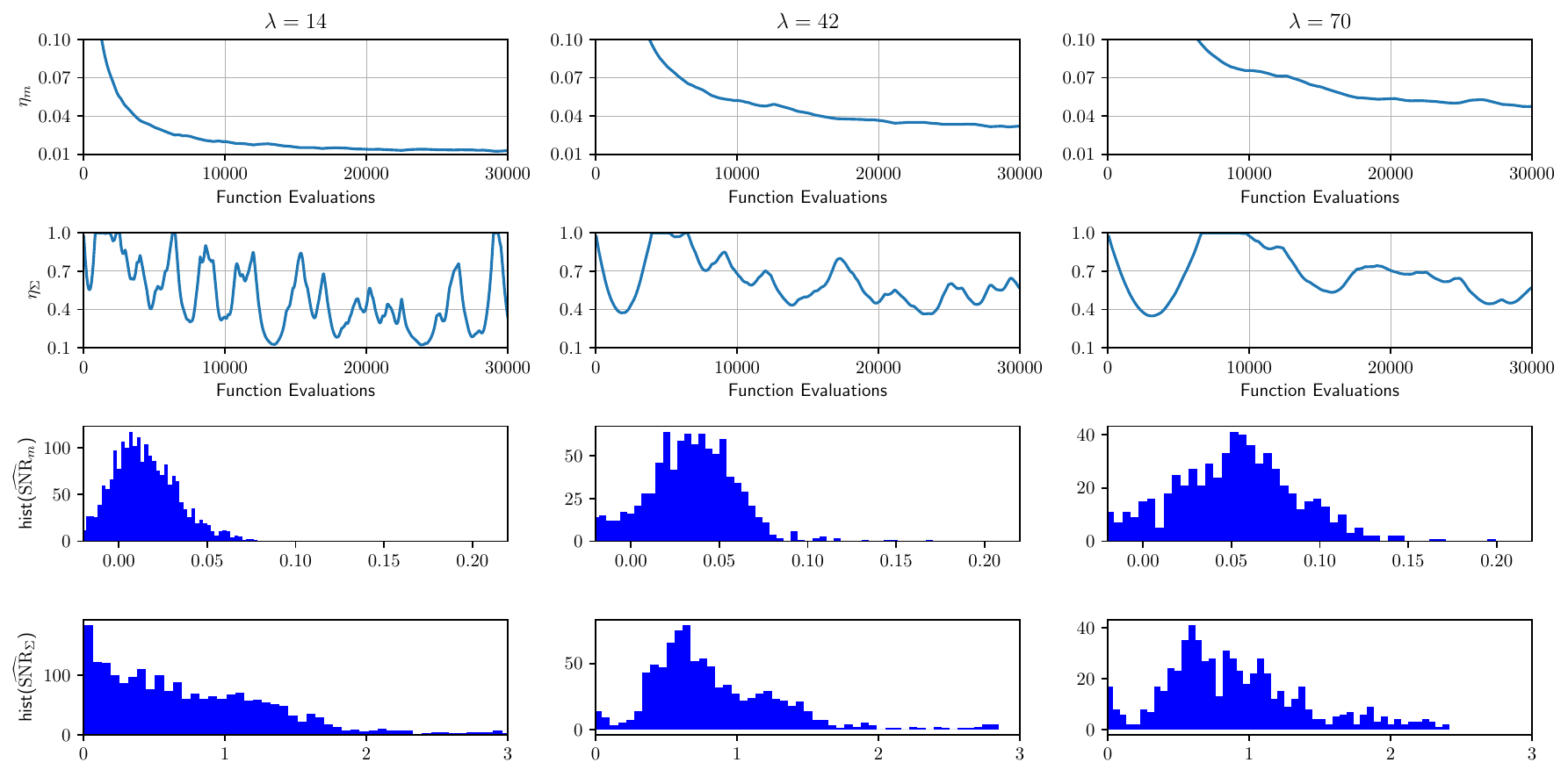}
  \caption{LRA-CMA-ES behaviors on the 30-D Sphere function with $\lambda \in \{ 14, 42, 70 \}$.
  $\eta_m$, $\eta_{\Sigma}$, and the histograms of the estimated SNR w.r.t. $m$ and $\Sigma$, in this order from the top.
  The y-axes in $\eta_m$ and $\eta_{\Sigma}$ are shown on the linear scale rather than the log scale.}
  \label{fig:adaptlr_behavior_popsize_learningrate}
\end{figure*}

Figure~\ref{fig:lamb_srsp1_d=30_small} shows the success rates and SP1 values with respect to $\lambda \in \{ 14, 28, 42, 56, 70 \}$ for the 30-D noiseless Sphere, Schaffer, and Rastrigin functions.
Although the SP1 value worsens with a larger $\lambda$ for the Rastrigin function, it appears to have a mild impact for the Sphere and Schaffer functions.
Figure~\ref{fig:adaptlr_behavior_popsize_learningrate} shows typical behaviors of LRA-CMA-ES with $\lambda \in \{14, 42, 70 \}$ on the 30-D Sphere function.
As $\lambda$ increases, it can be observed that the learning rates (especially $\eta_m$) also generally increase linearly.
This is because, as $\lambda$ increases, the SNR also increases, allowing for a larger learning rate to maintain the target SNR.
This phenomenon can also be theoretically explained as follows:
The SNR analysis for the infinite-dimensional Sphere function in Appendix~\ref{sec:app_discuss_snr} shows that under the assumption of the optimal step-size, $\mathrm{SNR} \approx O(\lambda/d)$.
In this case, increasing $\lambda$ can linearly increase the SNR;
therefore, it is expected that the learning rate can be kept linearly larger, which is consistent with our empirical findings.
However, this analysis was conducted using the (infinite-dimensional) Sphere function; thus, this discussion cannot be directly applied to multimodal problems.

\begin{figure*}[tb]
  \centering
  \includegraphics[width=0.964\hsize,trim=5 10 5 25,clip]{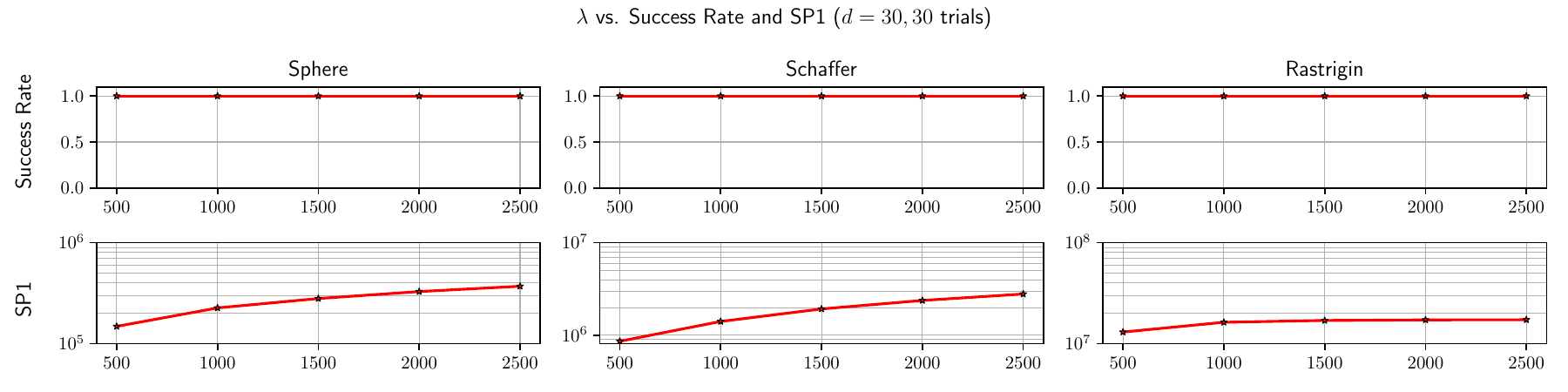}
  \caption{Success rates and SP1 values with $\lambda \in \{ 500, 1000, 1500, 2000, 2500 \}$ for 30-D noiseless problems (30 trials).}
  \label{fig:lamb_srsp1_d=30_large}
\end{figure*}

Additionally, we investigated the behavior for larger population sizes using various values of $\lambda \in \{ 500, 1000, 1500, 2000, 2500 \}$, as shown in Figure~\ref{fig:lamb_srsp1_d=30_large}.
Compared to the results for $\lambda \in \{ 14, 28, 42, 56, 70 \}$, the SP1 value remains almost constant for the Rastrigin function with respect to the $\lambda$ value.
However, in the Sphere and Schaffer functions, the SP1 value deteriorates slightly for larger $\lambda$ values.
This may be partially because the proposed method is designed to solve difficult problems (e.g., $\eta_m, \eta_{\Sigma} \leq 1$).
Although more aggressive learning rate updates may improve the performance, such strategies were beyond the scope of this study.

\subsection{LRA-CMA-ES vs. PSA-CMA-ES}
\label{sec:exp_vs_psa}
We compared the performance of the proposed LRA-CMA-ES with that of PSA-CMA-ES~\cite{psacmaes}, which is a state-of-the-art population size adaptation method.
For a fair comparison, we employed almost the same procedure and hyperparameters for PSA-CMA-ES as those for the CMA-ES described in Section~\ref{sec:cma}.
The only difference was that PSA-CMA-ES required additional normalization factors (Eqs.~(6) and (7) in ~\cite{psacmaes}) to derive an approximate value for the parameter movement.
For the step-size correction in PSA-CMA-ES, we used Blom's approximation to calculate the weighted average of the expected value of normal-order statistics~\cite{qualitygainwres}.
Additionally, we used the recommended values for the PSA-CMA-ES hyperparameters~\cite{psacmaes}.
The experimental settings were the same as those described in Section~\ref{sec:fixed_vs_adapt}.
All LRA-CMA-ES results were obtained from Section~\ref{sec:fixed_vs_adapt}.

\begin{figure*}[tb]
  \centering
  \includegraphics[width=0.964\hsize,trim=5 5 5 5,clip]{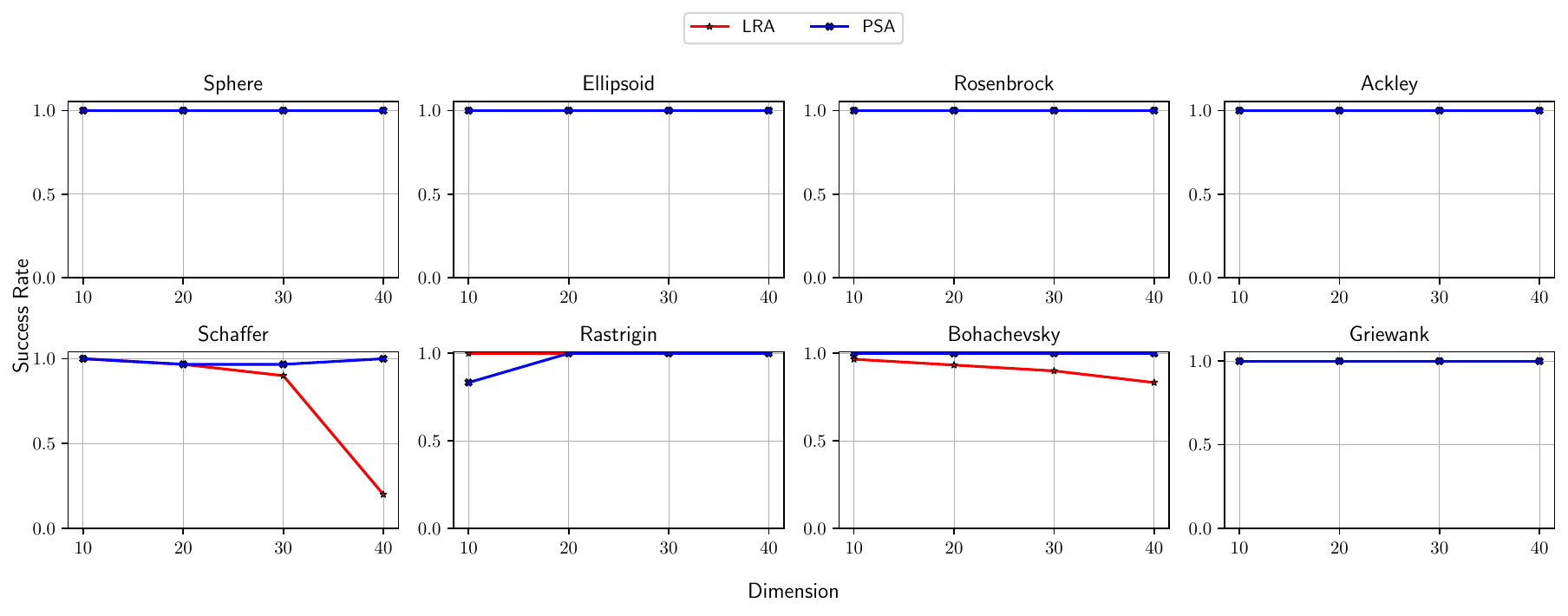}
  \caption{Performances of LRA-CMA-ES and PSA-CMA-ES: success rates according to the number of dimensions (noiseless problems).}
  \label{fig:psa_compare_succrate}
\end{figure*}

\begin{figure*}[tb]
  \centering
  \includegraphics[width=0.964\hsize,trim=5 5 5 5,clip]{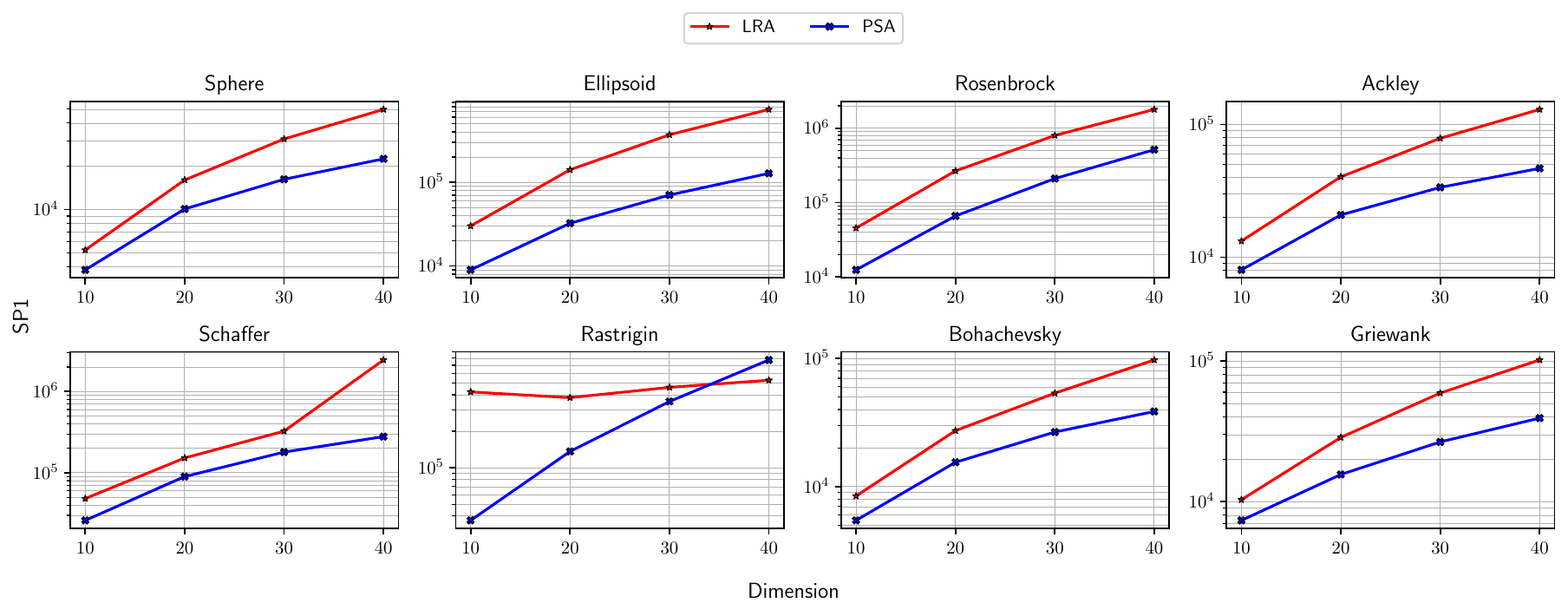}
  \caption{Performaces of LRA-CMA-ES and PSA-CMA-ES: SP1 values according to the number of dimensions (noiseless problems).}
  \label{fig:psa_compare_sp1}
\end{figure*}

\begin{figure*}[tb]
  \centering
  \includegraphics[width=0.964\hsize,trim=5 5 5 5,clip]{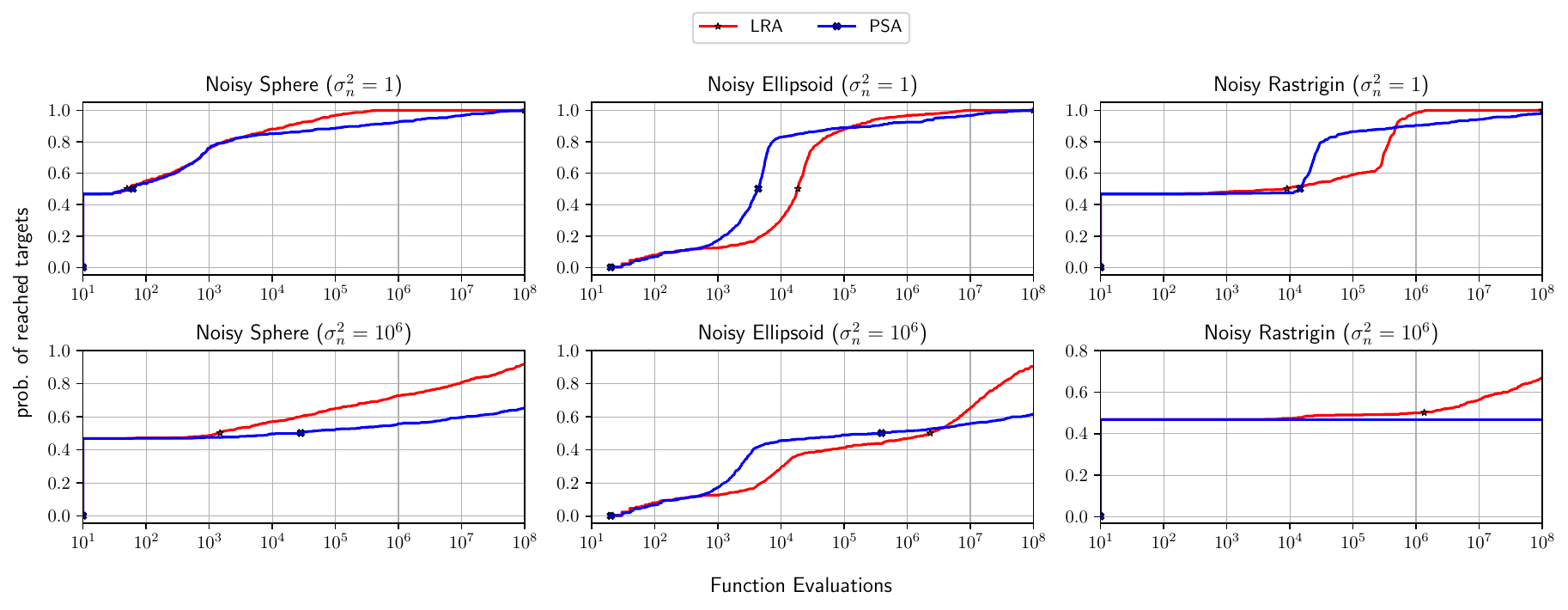}
  \caption{Performances of LRA-CMA-ES and PSA-CMA-ES: Empirical cumulative density function for 10-D noisy problems, with $\sigma_n^2$ set to $1$ or $10^6$.}
  \label{fig:psa_ecdf}
\end{figure*}

Figures~\ref{fig:psa_compare_succrate} and \ref{fig:psa_compare_sp1} show the success rates and SP1 values, respectively, for the noiseless problems,
wherein it is evident that PSA-CMA-ES exhibits better results than LRA-CMA-ES for most problems.
Figure~\ref{fig:psa_ecdf} illustrates the ECDF for noisy problems.
The performance of LRA-CMA-ES is better than that of PSA-CMA-ES in most of the tested cases.
For example, for the Rastrigin function with a strong-noise setting (bottom right of Figure~\ref{fig:psa_ecdf}), PSA-CMA-ES stopped improving the function value, whereas LRA-CMA-ES continued improving it.
These results suggest that LRA-CMA-ES and PSA-CMA-ES are suitable for different problems.
However, these performance differences can be mitigated to a certain degree by adjusting the hyperparameters of each method and do not necessarily suggest that there is a fundamental performance difference between learning rate and population size adaptations.
Although we still argue that LRA is more practically useful than population size adaptation, as described in Section~\ref{sec:intro}, a detailed comparison of these methods will be an interesting direction for future work.

\section{Conclusion}
\label{sec:conclusion}
This study presented the design principles and practices of LRA for CMA-ES.
We first demonstrated that difficult problems can be solved relatively easily by decreasing the learning rate and ensuring that the parameter behavior was closer to the ODE trajectory.
However, decreasing it excessively worsened the search efficiency. Therefore, we attempted to determine the optimal learning rate for maximizing the expected improvement, which was nearly proportional to the SNR under some assumptions.
Based on these observations, we developed a new LRA mechanism to solve multimodal and noisy problems using CMA-ES without extremely expensive learning-rate tuning.
The basic concept of the proposed algorithm, LRA-CMA-ES, is to adapt the learning rate such that the SNR can be kept constant, which is nearly optimal based on the optimal learning rate discussion.
Experiments involving noiseless multimodal problems revealed that the proposed LRA-CMA-ES can adapt the learning rate appropriately depending on the search situation, and it works well without tuning the learning rate.
Additionally, LRA-CMA-ES provided better solutions for noisy problems, even under strong-noise settings, which yielded problems that could not be solved by CMA-ES with a fixed learning rate.
In conclusion, LRA-CMA-ES effectively facilitates the solving of multimodal and noisy problems to a certain extent, eliminating the need for tuning the learning rate.

However, the proposed LRA-CMA-ES has some limitations, which will be addressed in future research.
First, it experienced several failures for the 40-D Schaffer function, although CMA-ES with an appropriately small learning rate succeeded with a high probability.
We believe that a detailed analysis of the SNR adaptation behavior is crucial to determine the reasons for this failure.
On a related note, our understanding of the appropriate hyperparameter settings in the proposed LRA mechanism remains limited.
Our experiments revealed that the hyperparameter settings affect the trade-off between stability and convergence speed.
Through experiment, we identified the hyperparameters that perform relatively well for noiseless and noisy problems; however, better configuration methods must be developed.
For example, the constant value $\mathcal{O}(1)$ is used for the cumulation factors $\beta_m$ and $\beta_{\Sigma}$; however, it may be more reasonable to consider that these factors depend on the parameter degrees of freedom, that is, $\beta_m = \mathcal{O}(1/d)$ and $\beta_{\Sigma} = \mathcal{O}(1/d^2)$.
In addition, the method for determining the appropriate value of the target SNR $\alpha$ can be refined further.
The SNR analysis presented in Appendix~\ref{sec:app_discuss_snr} implies that $\alpha = \mathcal{O}(\lambda / d)$ is reasonable for an infinite-dimensional Sphere function.
On the other hand, because this analysis cannot be directly applied to multimodal or noisy problems, there is still room to discuss the best method for determining $\alpha$.
A deeper understanding of the hyperparameter effects is crucial for improving the reliability of the proposed LRA-CMA-ES.

Additionally, LRA-CMA-ES, which mainly focuses on well-structured multimodal problems, alone cannot solve weakly structured ones, as discussed in Section~\ref{sec:intro}.
To address these situations, integrating restart strategies with LRA-CMA-ES is a promising direction.

Finally, developing a more rational LRA approach is an intriguing topic for future research.
Although our discussion in Section~\ref{sec:optimal_lr} offers valuable insights into designing an ideal learning rate, it was based on several assumptions and has the potential for improvement.
A more detailed theoretical study could result in a more rational design for learning rates, which is crucial for advancing this line of research.

\begin{acks}
This study was partially supported by JSPS KAKENHI, grant number 19H04179 and 23K11260.
\end{acks}


\bibliographystyle{ACM-Reference-Format}


\appendix

\section{Derivation for Section \ref{sec:snr_estimation}}
\label{sec:app_snr}

\subsection{Derivations of Eq.~(\ref{eq:mvae_dist})}
This section presents the detailed derivation of Eq.~(\ref{eq:mvae_dist}).
By ignoring $(1-\beta)^n$, $\mvae^{(t+n)}$ can be approximately calculated as follows:
\small
\begin{align*}
    \mvae^{(t+n)} &= (1 - \beta) \mvae^{(t+n-1)} + \beta \tilde\Delta^{(t+n-1)} \\
    &= (1-\beta) \left\{ (1-\beta) \mvae^{(t+n-2)} + \beta \tilde\Delta^{(t+n-2)} \right\} + \beta \tilde\Delta^{(t+n-1)} \\
    &= ...\\
    &= (1-\beta)^{n} \mvae^{(t)} + \sum_{i=0}^{n-1} (1-\beta)^i \beta \tilde\Delta^{(t+n-1-i)} \\
    &\approx \sum_{i=0}^{n-1} (1-\beta)^i \beta \tilde\Delta^{(t+n-1-i)}.
\end{align*}
\normalsize
Here, we assume the $\tilde\Delta^{(\cdot)}$ are uncorrelated with each other; this corresponds to the scenario where $\eta$ is sufficiently small.
In this case, we can ignore the dependence of $t$, that is, $\E[\tilde\Delta^{(t+n-1-i)}] =: \E[\tilde\Delta]$.
Thus,
\small
\begin{align*}
    \E[\mvae^{(t+n)}] = \sum_{i=0}^{n-1} (1-\beta)^i \beta \E[\tilde\Delta].
\end{align*}
\normalsize
where
\small
\begin{align*}
     \sum_{i=0}^{n-1} (1-\beta)^i &= \frac{1 \cdot \{ 1-(1-\beta)^{n} \} }{1-(1-\beta)} = \frac{1-(1-\beta)^{n}}{\beta}.
\end{align*}
\normalsize
Subsequently, ignoring $(1-\beta)^{n}$, we approximate $\E[\mvae^{(t+n)}]$ as
\small
\begin{align*}
    \E[\mvae^{(t+n)}] = [1-(1-\beta)^{n}] \E[\tilde\Delta] \approx \E[\tilde\Delta].
\end{align*}
\normalsize
Next, we consider the covariance $\Cov[\mvae^{(t+n)}]$:
\small
\begin{align*}
    \Cov[\mvae^{(t+n)}] = \E[\mvae^{(t+n)} (\mvae^{(t+n)})^{\top}] - \E[\mvae^{(t+n)}] (\E[\mvae^{(t+n)}])^{\top}.
\end{align*}
\normalsize
We first determine the exact expression for $\mvae^{(t+n)} (\mvae^{(t+n)})^{\top}$ as follows:
\small
\begin{align*}
    &\mvae^{(t+n)} (\mvae^{(t+n)})^{\top} = \beta^2 \sum_{i=0}^{n-1} (1-\beta)^{2i} \tilde\Delta^{(t+n-1-i)} (\tilde\Delta^{(t+n-1-i)})^{\top} \\
    &\quad + \beta^2 \sum_{i,j=0: i\neq j}^{n-1} (1-\beta)^i (1-\beta)^j \tilde\Delta^{(t+n-1-i)} (\tilde\Delta^{(t+n-1-j)})^{\top}.
\end{align*}
\normalsize
Note that, for $i,j \in \{0, \ldots , n-1 \} (i \neq j)$, $\E[\tilde\Delta^{(t+n-1-i)} (\tilde\Delta^{(t+n-1-j)})^{\top}]$ $= \E[\tilde\Delta] (\E[\tilde\Delta])^{\top}$ because we assume that they are not correlated.
For $i \in \{0, \ldots , n-1 \}$, $\E[\tilde\Delta^{(t+n-1-i)} (\tilde\Delta^{(t+n-1-i)})^{\top}] = \E[\tilde\Delta] (\E[\tilde\Delta])^{\top} + \Cov[\tilde\Delta]$.
Thus,
\small
\begin{align*}
    \E[\mvae^{(t+n)}& (\mvae^{(t+n)})^{\top}] \\
    &= \beta^2 \sum_{i=0}^{n-1} (1-\beta)^{2i} \left( \E[\tilde\Delta] (\E[\tilde\Delta])^{\top} + \Cov[\tilde\Delta] \right) \\
    &\quad + \beta^2 \sum_{i,j=0: i\neq j}^{n-1} (1-\beta)^i (1-\beta)^j \E[\tilde\Delta] (\E[\tilde\Delta])^{\top}, \\
    &= \E[\mvae^{(t+n)}]  (\E[\mvae^{(t+n)}])^{\top} + \beta^2 \sum_{i=0}^{n-1} (1-\beta)^{2i} \Cov[\tilde\Delta].
\end{align*}
\normalsize
Therefore,
\small
\begin{align*}
    \Cov[\mvae^{(t+n)}] &= \E[\mvae^{(t+n)} (\mvae^{(t+n)})^{\top}] - \E[\mvae^{(t+n)}] (\E[\mvae^{(t+n)}])^{\top} \\
    &= \beta^2 \sum_{i=0}^{n-1} (1-\beta)^{2i} \Cov[\tilde\Delta].
\end{align*}
\normalsize
Here,
\small
\begin{align*}
    \sum_{i=0}^{n-1} (1-\beta)^{2i} = \frac{1-(1-\beta)^{2n}}{1-(1-\beta)^2} = \frac{1-(1-\beta)^{2n}}{\beta(2-\beta)}.
\end{align*}
\normalsize
Thus, by ignoring $(1-\beta)^{2n}$, $\Cov[\mvae^{(t+n)}]$ can be approximated as
\small
\begin{align*}
    \Cov[\mvae^{(t+n)}] &= [1-(1-\beta)^{2n}] \frac{\beta}{2-\beta} \Cov[\tilde\Delta], \\
    &\approx \frac{\beta}{2-\beta} \Cov[\tilde\Delta].
\end{align*}
\normalsize
Therefore, $\mvae^{(t+n)}$ approximately follows the following distribution:
\small
\begin{align*}
    \mvae^{(t+n)} 
    \sim \mathcal{D}\left( \E[\tilde\Delta], \frac{\beta}{2 - \beta} \Cov[\tilde\Delta]\right) .
\end{align*}
\normalsize
Thus, the derivation of Eq.~(\ref{eq:mvae_dist}) is complete.

\subsection{Derivation of Estimates for $\| \E[\tilde\Delta] \|_2^2$}
We organized the relation between $\mvae$ and $\tilde\Delta$ using the following equation:
\begin{align*}
\small
    \E[\| \mvae \|_2^2] &= \E[\mvae]^{\top} I \E[\mvae] + \Tr(\Cov[\mvae]) \nonumber \\
    &\approx \| \E[\tilde\Delta] \|_2^2 + \Tr \left( \frac{\beta}{2-\beta} \Cov[\tilde\Delta] \right) \nonumber  \\
    &= \| \E[\tilde\Delta] \|_2^2 + \frac{\beta}{2-\beta} \Tr(\Cov[\tilde\Delta]).
\end{align*}
\normalsize

Now, we apply the same arguments to $\mvav$ and obtain:
\begin{align*}
\small
    \E[\mvav] &= [1-(1-\beta)^{t+1}] \E[ \| \tilde\Delta \|_2^2 ] \nonumber \\
    &\approx \E[\| \tilde\Delta \|_2^2] = \| \E[\tilde\Delta] \|_2^2 + \Tr(\Cov[\tilde\Delta]).
\end{align*}
\normalsize
By reorganizing these arguments, we obtain
\begin{align*}
\small
    \| \E[\tilde\Delta] \|_2^2 \approx \frac{2-\beta}{2-2\beta} \E[\| \mvae \|_2^2] - \frac{\beta}{2-2\beta} \E[ \mvav ].
\end{align*}
\normalsize
This provides the rationale for estimating $\frac{2-\beta}{2 - 2\beta} \norm{\mvae}_2^2 - \frac{\beta}{2-2\beta} \mvav$ for $\| \E[\tilde\Delta] \|_2^2$.

\section{Theoretical and Empirical Insights into SNR}
\label{sec:app_discuss_snr}

In Section~\ref{sec:lra}, we assumed that the signal-to-noise ratio (SNR) is relatively small, for example, $\mathrm{SNR} \lessapprox 1$, which validates the approximation $1 / (1 + \mathrm{SNR}^{-1}) \approx \mathrm{SNR}$.
In this section, we theoretically and empirically discuss the validity of $\mathrm{SNR} \lessapprox 1$.

To obtain useful insights into this $\mathrm{SNR}$ from a theoretical perspective, we considered observing it in a situation wherein the objective function is the sphere function $f(x) = \norm{x}^2$ and the covariance matrix is $\Sigma = \sigma^2 I$, where $\sigma = \bar{\sigma} \frac{\norm{m}}{d}$ and $\bar{\sigma}$ is called the normalized step-size. 
The quality gain analysis \cite{opt_wr,qualitygainwres} implies that for a sufficiently large $d$, the distribution of the $i$th ranked solution among the $\lambda$ candidate solutions is approximated as $X_{i:\lambda} = m + \sigma \mathcal{N}_{i:\lambda}\frac{m}{\norm{m}} + \sigma \mathcal{N}_{i}^{\bot}$, where $\mathcal{N}_{i:\lambda}$ is the $i$th order statistics among $\lambda$ normally distributed random variables and $\mathcal{N}_{i}^{\bot}$ is an independently distributed $d$ dimensional normal random vector with covariance matrix $I - \frac{m m^\T}{\norm{m}^2}$ if $m \neq 0$. 
Using this approximation, we obtain $\Delta_m = \sigma \left(\sum_{i=1}^{\lambda} w_i \mathcal{N}_{i:\lambda}\right) \frac{m}{\norm{m}} + \sigma \left(\sum_{i=1}^{\lambda} w_i \mathcal{N}_{i}^{\bot}\right)$. 
Let $\bm{w} = (w_1, \dots, w_\lambda)$, $\bm{n}_{(\lambda)} = (\E[\mathcal{N}_{1:\lambda}], \dots, \E[\mathcal{N}_{\lambda:\lambda}])$.
$\bm{N}_{(\lambda)}$ is a matrix whose $(i,j)$th element is $\E[\mathcal{N}_{i:\lambda}\mathcal{N}_{j:\lambda}]$.
Then, we obtain
\begin{subequations}
\begin{align}
    \E[\Delta_m] &= \sigma (\bm{w}^\T \bm{n}_{(\lambda)}) \frac{m}{\norm{m}}, \\
    \E[\Delta_m \Delta_m^\T] &= \sigma^2 (\bm{w}^\T \bm{N}_{(\lambda)} \bm{w}) \frac{m m^\T}{\norm{m}^2} + \sigma^2 \norm{\bm{w}}^2 \left(I - \frac{m m^\T}{\norm{m}^2}\right).
\end{align}
\end{subequations}
Because $F_m = \sigma^{-2} I$, we obtain
\begin{subequations}
\begin{align}
    \mathrm{SNR} 
    &= \frac{  \sigma^{-2} \norm{\E[\Delta_m]}^2 }{ \sigma^{-2} \Tr(\E[\Delta_m \Delta_m^\T]) - \sigma^{-2} \norm{\E[\Delta_m]}^2 } \\
    &= \frac{(\bm{w}^\T \bm{n}_{(\lambda)})^2}{ \bm{w}^\T \bm{N}_{(\lambda)} \bm{w} + (d-1) \norm{\bm{w}}^2 - (\bm{w}^\T \bm{n}_{(\lambda)})^2} \\
    &\approx \frac{(\bm{w}^\T \bm{n}_{(\lambda)})^2}{ (d-1) \norm{\bm{w}}^2}\\
    &= \frac{1}{d-1} \frac{(\bm{w}^\T \bm{n}_{(\lambda)})^2}{ \norm{\bm{w}}^2 }\\
    &\approx \frac{\lambda}{d-1} \frac{(\bm{w}^\T \bm{n}_{(\lambda)})^2}{ \norm{\bm{w}}^2 \norm{\bm{n}_{(\lambda)}}^2 }.
\end{align}
\end{subequations}
Here, we used the following asymptotically true approximations for $\lambda$ (See Eq.~(A2) provided in \cite{qualitygainwres}):
\begin{align}
    \frac{ \bm{w}^\T \bm{N}_{(\lambda)} \bm{w} }{ (\bm{w}^\T \bm{n}_{(\lambda)})^2 } \approx 1 
    &&\text{and}
    &&
    \frac{ \norm{\bm{n}_{(\lambda)}}^2 }{ \lambda } \approx 1.
\end{align}
It should be noted that $\frac{(\bm{w}^\T \bm{n}_{(\lambda)})^2}{ \norm{\bm{w}}^2 \norm{\bm{n}}^2 }$ is upper bounded by $0.25$ if only non-negative weights are used for the $m$-update, which aligns with our weight scheme.
Therefore, we can expect that $\mathrm{SNR} \lessapprox 1$ holds if $\lambda$ is not considerably large relative to $d$; for example, $\lambda \leq 4(d-1)$.
Importantly, in difficult problems, such as multimodal problems, the SNR tends to be smaller than that in the sphere functions. Therefore, it should be noted that the assumption of $\mathrm{SNR} \lesssim 1$ becomes more easily valid for such difficult problems.

\begin{figure*}[tb]
  \centering
  \includegraphics[width=0.964\hsize,trim=5 5 5 5,clip]{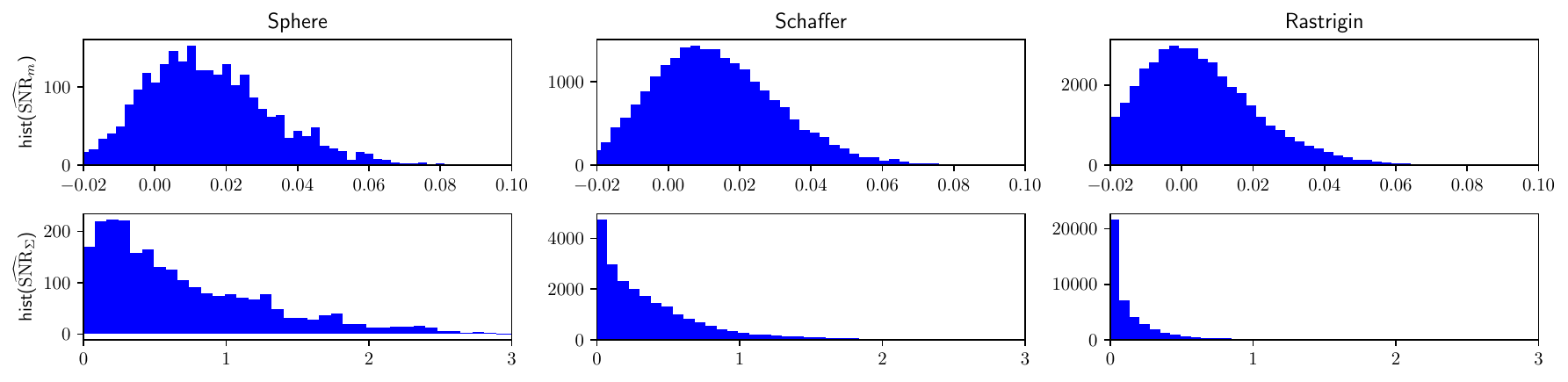}
  \caption{Histogram of the estimated SNR in typical trials on 30-D noiseless problems. Estimated SNR with respect to (top) the mean vector $m$ and (bottom) the covariance matrix $\Sigma$. The SNR was estimated using the method described in Section~\ref{sec:snr_estimation}.}
  \label{fig:adaptlr_behavior_snr}
\end{figure*}

The main limitation of the aforementioned analysis is the assumption that the dimension $d$ and the population size $\lambda$ are sufficiently large.
To verify whether the assumption $\mathrm{SNR} \lessapprox 1$ works in practice, we conducted experiments using the LRA-CMA-ES for $30$-dimensional Sphere, Schaffer, and Rastrigin functions using the same settings as those mentioned in Section~\ref{sec:setup}, and $\lambda = 14$ for $d = 30$.
Figure~\ref{fig:adaptlr_behavior_snr} illustrates the typical behavior of the estimated SNR where it was estimated using the method described in Section~\ref{sec:snr_estimation}.
It should be noted that this value includes estimation errors. 
Although the estimated SNR for the covariance in the Sphere function tends to be slightly larger, it often remains under $1$, particularly for more \emph{difficult} problems such as the Rastrigin function.
These results suggest that the assumption of SNR to be small, e.g., $\mathrm{SNR} \lessapprox 1$, appears to be valid to a certain degree even under finite dimensions and population sizes.

\section{Additional Experimental Results}
\label{sec:app_expresults}

\begin{figure*}[tb]
  \centering
  \includegraphics[width=0.964\hsize,trim=5 5 5 5,clip]{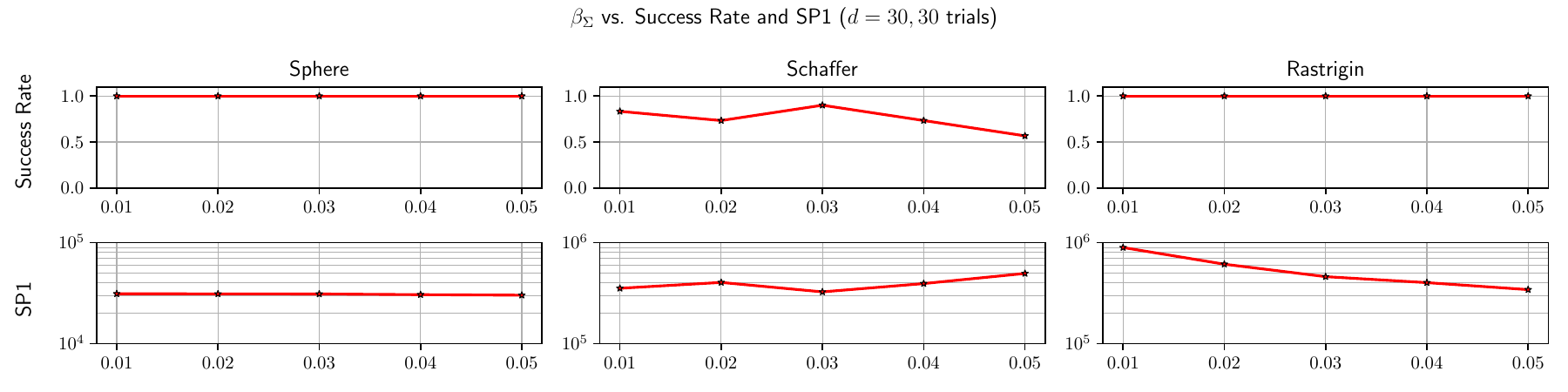}
  \caption{Success rate and SP1 values with hyperparameter $\beta_{\Sigma} \in \{ 0.01, 0.02, ..., 0.05 \}$  on 30-D noiseless problems.}
  \label{fig:beta_Sigma_srsp1_d=30_detailed}
\end{figure*}

Figure~\ref{fig:beta_Sigma_srsp1_d=30_detailed} shows the success rate and SP1 values with respect to $\beta_{\Sigma} \in \{ 0.01, 0.02, ..., 0.05 \}$ for the 30-D noiseless Sphere, Schaffer, and Rastrigin functions.
Clearly, the performance is not significantly affected by $\beta_{\Sigma}$ values within this range. However, as shown in Figure~\ref{fig:beta_Sigma_srsp1_d=30}, an excessively small $\beta_{\Sigma}$ value decelerates the convergence for the Rastrigin function.

\begin{figure*}[tb]
  \centering
  \includegraphics[width=0.964\hsize,trim=5 5 5 5,clip]{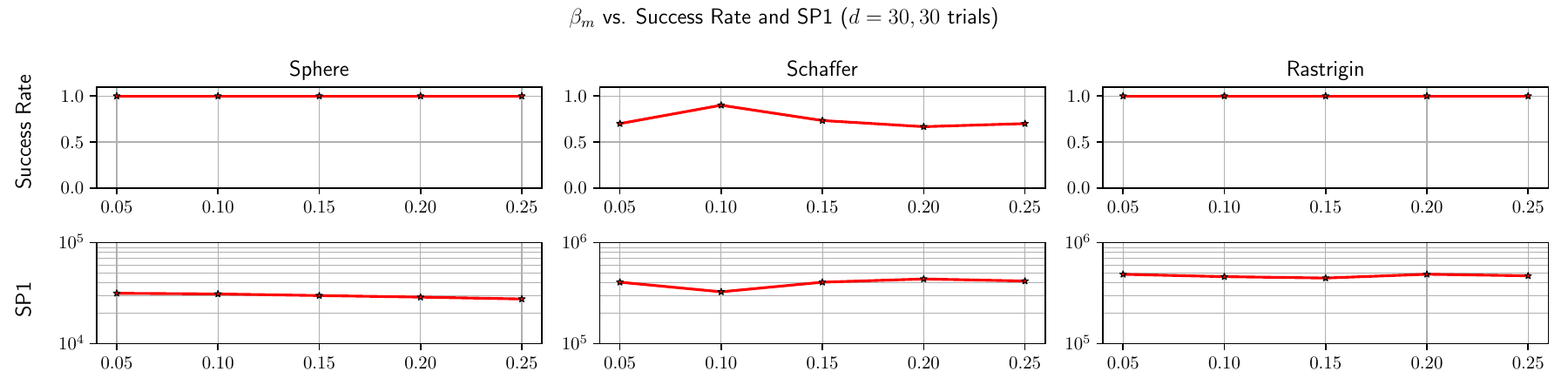}
  \caption{Success rate and SP1 values with hyperparameter $\beta_m$ for 30-D noiseless problems.}
  \label{fig:beta_mean_srsp1_d=30}
\end{figure*}

\begin{figure*}[htb]
  \centering
  \includegraphics[width=0.964\hsize,trim=5 5 5 5,clip]{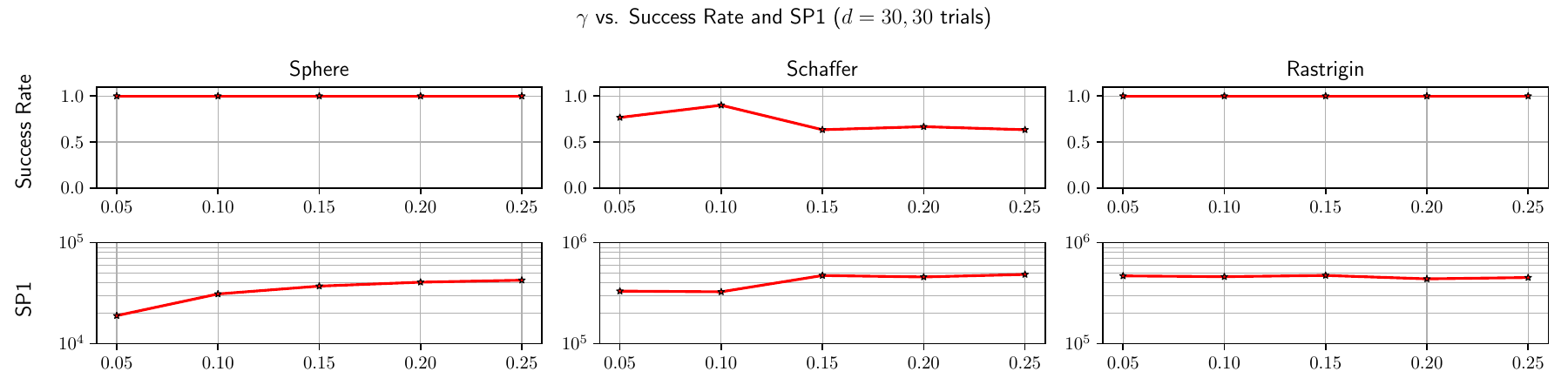}
  \caption{Success rate and SP1 values with hyperparameter $\gamma$ for 30-D noiseless problems.}
  \label{fig:gamma_srsp1_d=30}
\end{figure*}

Figures~\ref{fig:beta_mean_srsp1_d=30} and \ref{fig:gamma_srsp1_d=30} show the success rates and SP1 values for $\beta_m$ and $\gamma$, respectively.
The results show that the performance is relatively stable against these hyperparameters.

\end{document}